\documentclass{article}

\usepackage{amsmath,amsfonts,bm}

\def\eqref#1{equation~\ref{#1}}

\def\1{\bm{1}}

\def\vb{{\bm{b}}}

\def\vw{{\bm{w}}}
\def\vx{{\bm{x}}}

\def\vphi{{\bm{\phi}}}

\def\mA{{\bm{A}}}

\def\mC{{\bm{C}}}
\def\mD{{\bm{D}}}

\def\mH{{\bm{H}}}
\def\mI{{\bm{I}}}
\def\mJ{{\bm{J}}}
\def\mK{{\bm{K}}}

\def\mM{{\bm{M}}}

\def\mP{{\bm{P}}}

\def\mT{{\bm{T}}}
\def\mU{{\bm{U}}}
\def\mV{{\bm{V}}}
\def\mW{{\bm{W}}}
\def\mX{{\bm{X}}}

\def\mZ{{\bm{Z}}}

\def\mSigma{{\bm{\Sigma}}}

\DeclareMathAlphabet{\mathsfit}{\encodingdefault}{\sfdefault}{m}{sl}
\SetMathAlphabet{\mathsfit}{bold}{\encodingdefault}{\sfdefault}{bx}{n}

\DeclareMathOperator*{\argmin}{arg\,min}

\usepackage[preprint]{neurips_2026}

\usepackage[utf8]{inputenc} 
\usepackage[T1]{fontenc}    
\usepackage{hyperref}       
\usepackage{url}            
\usepackage{booktabs}       
\usepackage{amsfonts}       
\usepackage{nicefrac}       
\usepackage{microtype}      
\usepackage[dvipsnames]{xcolor}         
\usepackage{lipsum}
\usepackage{amsmath}
\usepackage{amssymb}
\usepackage{graphicx}
\usepackage[capitalise, noabbrev, nameinlink]{cleveref}
\usepackage{xspace}
\usepackage{bbm}
\usepackage{algorithm}
\usepackage{algorithmic}
\usepackage{wrapfig}
\hypersetup{
    colorlinks=true,
    linkcolor=Sepia,
    citecolor=Sepia,
    urlcolor=Sepia,
}
\usepackage{placeins}
\usepackage{duckuments}
\usepackage{caption}
\usepackage{titlesec}
\usepackage{multirow}
\usepackage{fontawesome5}
\usepackage{hyperref}

\makeatletter
\newcommand{\githublink}{%
  \if@preprint
    \href{https://github.com/niniack/dmd\_vit}{\faGithub\ \texttt{https://github.com/niniack/dmd\_vit}}%
  \fi
}
\makeatother

\DeclareMathOperator{\cossim}{cossim}

\newcommand{\fullDMD}{full DMD\xspace}
\newcommand{\FullDMD}{Full DMD\xspace}
\newcommand{\anchoredDMD}{anchored DMD\xspace}

\titlespacing{\section}{0pt}{0.4\parskip}{0.4\parskip}
\titlespacing*{\paragraph}{0pt}{*0.5}{1em}

\title{Dynamic Mode Decomposition \\ along Depth in Vision Transformers}

\author{%
  Nishant Suresh Aswani \\
  NYU Tandon \& NYU Abu Dhabi\\
  \texttt{nishantaswani@nyu.edu} \\
  \And
  Saif Eddin Jabari \\
  NYU Tandon \& NYU Abu Dhabi\\
  \texttt{sej7@nyu.edu} \\
}

\begin{document}

\maketitle

\vspace{-1em}
\centerline{\githublink}
\vspace{1em}

\begin{abstract}
Recent work has shown that contiguous vision transformer (ViT) blocks (a) can be replaced by a linear map and (b) organize into recurrent phases of computation. We ask whether these observations coincide: does ViT depth implement approximately \textit{autonomous linear} dynamics, admitting a single operator \(\mK\) applied recurrently across a contiguous span? We test this using Dynamic Mode Decomposition (DMD), which fits \(\mK\) from selected, consecutive hidden-state pairs and predicts \(p\) steps ahead via \(\mK^p\). On four pretrained DINO ViTs, we study the regularization, rank, and calibration budget required for stable fitting. For short spans (\(p \leq 4\)), \(\mK^p\) tracks an unconstrained endpoint map to within \(0.02\) cosine similarity on DINOv3-H/16+, while also recovering intermediate activations at each skipped block. At early cut starts, the fitted operators compress to rank \(\ll d\) with minimal calibration data, and across tokens, \texttt{cls} is most amenable to linearization; both properties decay monotonically with depth. Yet this local fidelity does not transfer downstream. At the final hidden state, after propagating through the remaining blocks, an identity baseline becomes competitive.
\end{abstract}

\section{Introduction}
\label{sec:intro}

Recently, \citet{jacobs_block_2026} demonstrated that Vision Transformers (ViTs) organize into a small number of ``contiguous phases of computation'', i.e.\ consecutive transformer blocks can be approximated by a single transformer block applied recurrently. In parallel, \citet{shopkhoev_replaceme_2025} show that a span of blocks can be substituted with a linear transformation, effectively skipping over a chunk of non-linear computation with a linear map. Motivated by these observations, we study whether the recurrent structure observed by \citet{jacobs_block_2026} and the linear replacement suggested by \citet{shopkhoev_replaceme_2025} intersect in a middle ground:

\textit{Can a span of transformer blocks be replaced by the recurrent application of a single linear operator?}

\begin{figure}[!htbp]
    \centering
    \includegraphics[width=\textwidth]{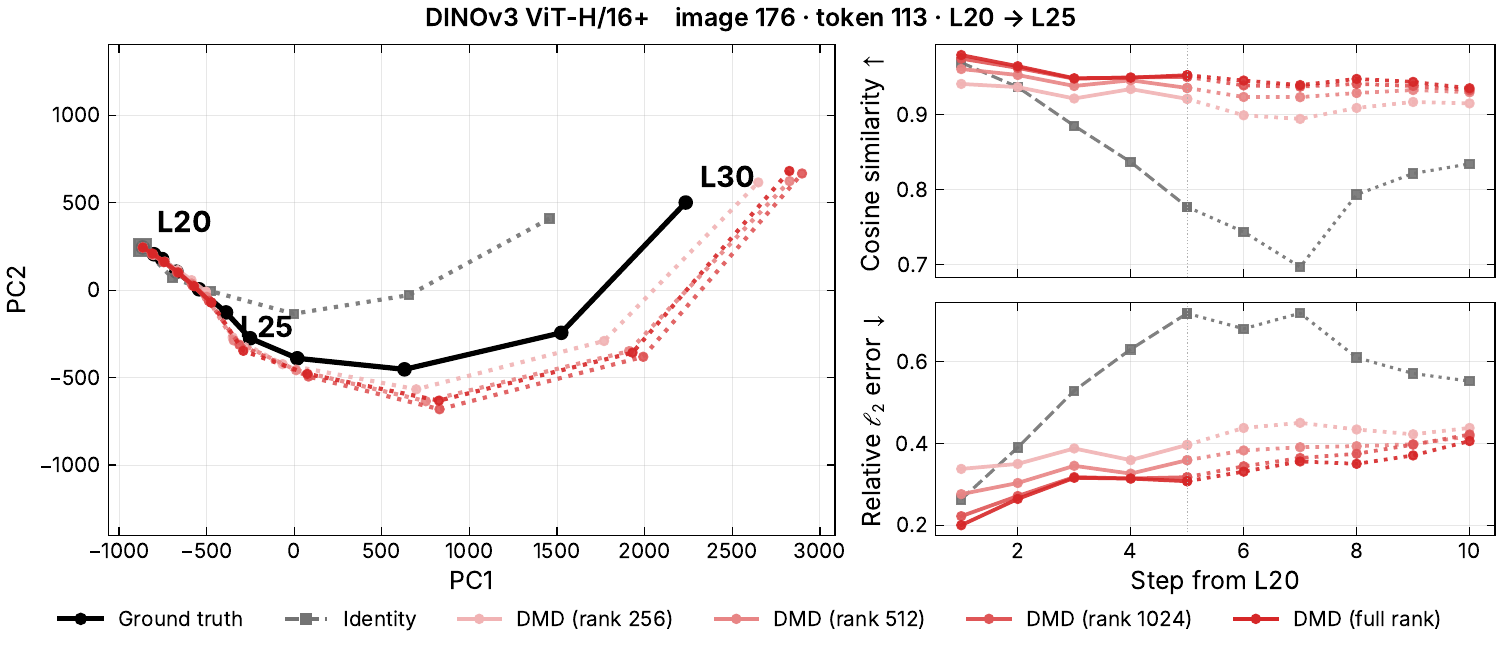}
    \caption{\textbf{DMD approximates a span of transformer blocks with a single recurrent linear map.} A single image-token pair from DINOv3-H/16+ at cut start \(i{=}20\) and prune length \(p{=}5\). The left panel shows the hidden-state trajectory across the spanned blocks in PCA space, comparing the ground truth to \fullDMD predictions \(\mK^q \mX_i\) at progressively higher ranks (\(r \in \{256, 512, 1024, d\}\)) and to an identity baseline. The right panels show cosine similarity (top) and relative \(\ell_2\) error (bottom) of the same predictions at each step \(q\) within the span.}
    \label{fig:hero}
    \vspace{-1em}
\end{figure}

To study this question, we employ the dynamic mode decomposition (DMD) \citep{schmid_dynamic_2010} algorithm, a core tool in the study of dynamical systems, to fit a linear map that, applied recurrently, approximates the hidden states across a selected span of blocks. \cref{fig:hero} provides a preview of our approach, where we plot the ground truth trajectory alongside the linear map's trajectory, in principal component space. Overall, in this work:

\begin{itemize}
  \item We introduce two DMD-based formulations to replace a span of transformer blocks with the recurrent application of a linear map.
  
  \item We characterize the design space (architecture, training data budget, and rank selection), discussing how these choices affect the quality of the operator on intermediate and downstream representations
\end{itemize}

\paragraph{Linearity and recurrence in depth.}
The interpretation of neural network depth as discrete time in a dynamical system, proposed by \citet{e_proposal_2017}, was later operationalized by \citet{chen_neural_2018, haber_stable_2017}. Within this perspective, we focus on a maturing body of work which observes that these `depth dynamics' may be approximately modeled as linear. \citet{li_residual_2023} compute the Jacobians of residual networks and find that the top singular vectors of neighboring Jacobians `align' across depth. \citet{aubry_transformer_2025a} extend this analysis to transformer-based models, uncovering an analogous alignment across both depth and tokens. Based on this, they hypothesize that representations may be modeled as a linear ODE. In parallel, \citet{aswani_koopman_2025} demonstrate that a Koopman autoencoder can model the layer-to-layer evolution of a simple residual model as a linear dynamical system in a \textit{learned}, latent space. Along a different thread, \citet{razzhigaev_your_2024} report near-perfect Procrustes similarity (\(\sim\!0.99\)) between two adjacent layer activations across several transformer models. Building on this, \citet{shopkhoev_replaceme_2025} show that a span of LLM blocks can be replaced by a single linear map with minimal loss in downstream performance. The latter findings establish that individual or multi-block transitions are well-described by a linear map, but leave open whether consecutive single-step transitions share a common map. Separately, \citet{jacobs_block_2026} present the Block-Recurrent Hypothesis (BRH), positing that a ViT's \(L\) blocks decompose into \(k \ll L\) contiguous phases, each re-writable with a single (non-linear) block applied recurrently. Whether this recurrent structure admits a purely \textit{linear} description remains untested. Our work tests that question.

\paragraph{DMD and neural networks.}
DMD \citep{schmid_dynamic_2010} extracts the best-fit linear operator from sequential data. In the context of deep learning, earlier work \citep{dogra_optimizing_2020, mohr_applications_2021} has applied DMD to parameter trajectories over gradient-descent training iterations, targeting optimization dynamics. Our focus is on depth dynamics. More relevant to our work, \citet{jacobs_block_2026} fit DMD to pre-processed ViT hidden states to analyze the `directional' dynamics. Their use of DMD is, however, restricted to studying the spectral properties on preprocessed activations, not for the goal of layer replacement. The works closest to ours do take this step. \citet{sugishita_extraction_2024} and \citet{aswani_representing_2025} both apply DMD along network depth to substitute nonlinear layers with linear maps, using monomial and Gaussian RBF dictionaries, and time-delay embeddings as observables (embedding function), respectively. Although, both operate on small-scale MLPs and work in lifted observable spaces. We depart from these by applying DMD without nonlinear observables, and at foundation-model scale.

\paragraph{Depth pruning.}
Structured depth pruning removes entire blocks to reduce cost. ShortGPT \citep{men_shortgpt_2025} ranks layers by a `Block Influence score' (one minus the mean cosine similarity between a block's input and output) and drops the lowest-scoring ones. \citet{gromov_unreasonable_2024} similarly prune a contiguous block of later layers identified by angular distance, relying on QLoRA healing to recover performance. \citet{shopkhoev_replaceme_2025} take a different approach, replacing the pruned span with a single least-squares linear map. Our study shares this paradigm of fitting-based linear replacement but differs in what is optimized. ReplaceMe fits an unconstrained endpoint map \(\mT\) from boundary activations only (shown in \cref{eq:rm-obj}). In this study, we fit a one-step operator \(\mK\) from intermediate states and enforce the composition \(\mK^n\) at prediction time (\cref{eq:full_dmd}). Comparing \(\mK^n\) against \(\mT\) isolates the time-invariance hypothesis. Agreement indicates approximately autonomous linear dynamics within the span, while a gap indicates depth-dependent structure that a single recurrence cannot capture. Given the shared ideas, we adopt similar terms from this literature, namely cut start and prune length, referring to the layer location where block replacement begins and the number of blocks being replaced, respectively.

In all of our experiments below, we use the two deepest, register-token variants of the DINOv2 (ViT-L/14 and ViT-G/14) models \citep{oquab_dinov2_2023, darcet_vision_2024}, as well as the two deepest DINOv3 (ViT-L/16 and ViT-H/16+) models \citep{simeoni_dinov3_2025}; the latter include register tokens by default. However, when fitting linear operators, we always discard the four register tokens, as they pollute the dynamics. We provide experiment details in \cref{app:sweep_details}. 
\section{Extrapolating One-Step Linear Maps}
\label{sec:motivation}

As summarized in \cref{sec:intro}, prior work, in particular that of \citet{razzhigaev_your_2024} and \citet{shopkhoev_replaceme_2025}, establishes that (at least) single blocks in large pre-trained models can be modeled with a linear map. This motivates a subsequent question.

\textit{If we fit a linear map \(\mT\) to replace a single transformer block, how many subsequent transformer blocks can \(\mT\) extrapolate?}

\begin{figure}[!htbp]
    \centering
    \includegraphics[width=\textwidth]{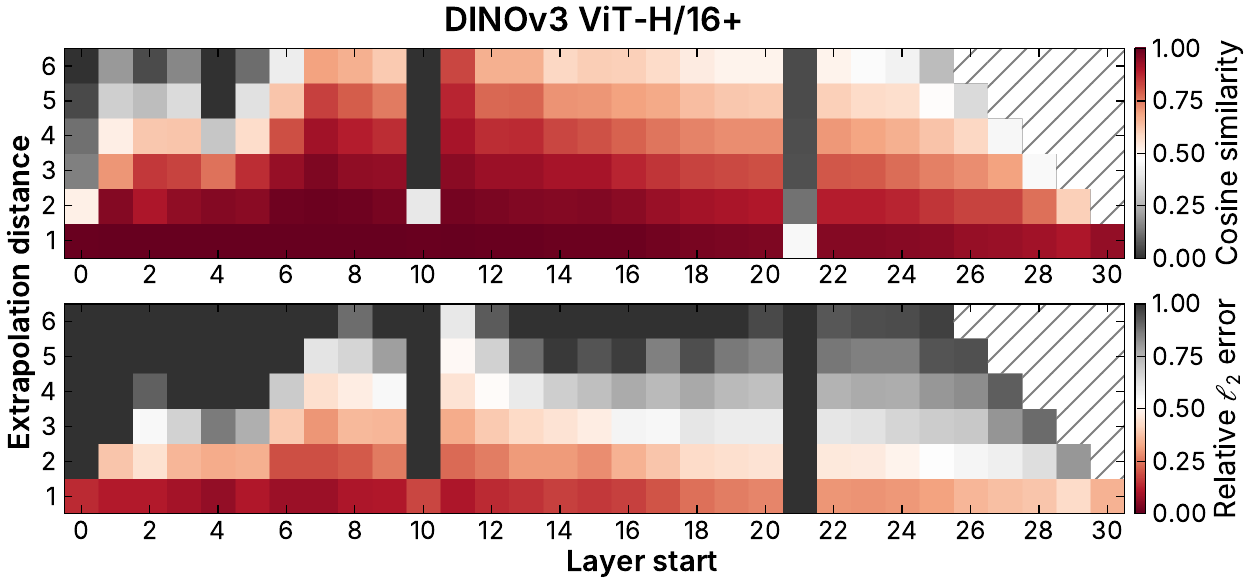}
    \caption{\textbf{Multi-step extrapolation of a one-step linear map.} A linear operator \(\mT_i\) is estimated via least squares to predict \(\mX_{i+1}\), and the \(n\)-th matrix power \(\mT_i^n\) is evaluated against the true activation \(\mX_{i+n}\). The horizontal axis is the starting layer \(i\) and the vertical axis is the extrapolation distance \(n\). The top row reports cosine similarity and the bottom row reports relative \(\ell_2\) error, capped at \(1.0\) for visual clarity. In the uncapped data, failed spans reach relative errors exceeding \(10^6\).}
    \label{fig:extrap}
    \vspace{-1em}
\end{figure}
For each layer \(i\) of DINOv3-H/16+ \citep{simeoni_dinov3_2025}, we fit a one-step operator \(\mT_i\) approximating \(\mX_i \mapsto \mX_{i+1}\), and evaluate the \(n\)-th matrix power \(\mT_i^n\) against the true \(\mX_{i+n}\) for \(n \in \{1, \ldots, 10\}\). We report mean cosine similarity (directional agreement) and mean relative \(\ell_2\) error (norm divergence), defined in \cref{app:prelim}. The one-step fit (\cref{fig:extrap}, top row, \(n{=}1\)) is strong at every layer except 21, with \(\cossim > 0.89\), validating the per-layer linear approximation of \citet{razzhigaev_your_2024}. Multi-step extrapolation, however, depends sharply on the starting layer. From layer 11, reusing \(\mT_i\) for five steps still yields \(\cossim \approx 0.87\), whereas from layer 10 it collapses to \(\cossim \approx -0.45\). The relative \(\ell_2\) error (bottom row) agrees. At layers 0, 10, and 21 it grows several orders of magnitude within a few steps. Spans starting at layers 6--9 and 11 are relatively better, retaining meaningful directional agreement and bounded error across several steps. \cref{fig:extrap-all} shows that larger models sustain faithful extrapolation over longer spans. While the failure of multi-step extrapolation is unsurprising, given that depth-wise dynamics need not be autonomous, the limited success at mid-depth spans warrants exploration. This motivates the \(\mK^n\) model formalized in \cref{sec:method}.


\newpage
\section{The Linear and Autonomous Hypothesis}
\label{sec:method}
We now turn to the central question of this work: \textit{can we replace transformer blocks with a recurrently applied linear operator?}

\begin{wrapfigure}[23]{r}{0.6\textwidth}
    \centering
    \includegraphics[width=0.6\textwidth]{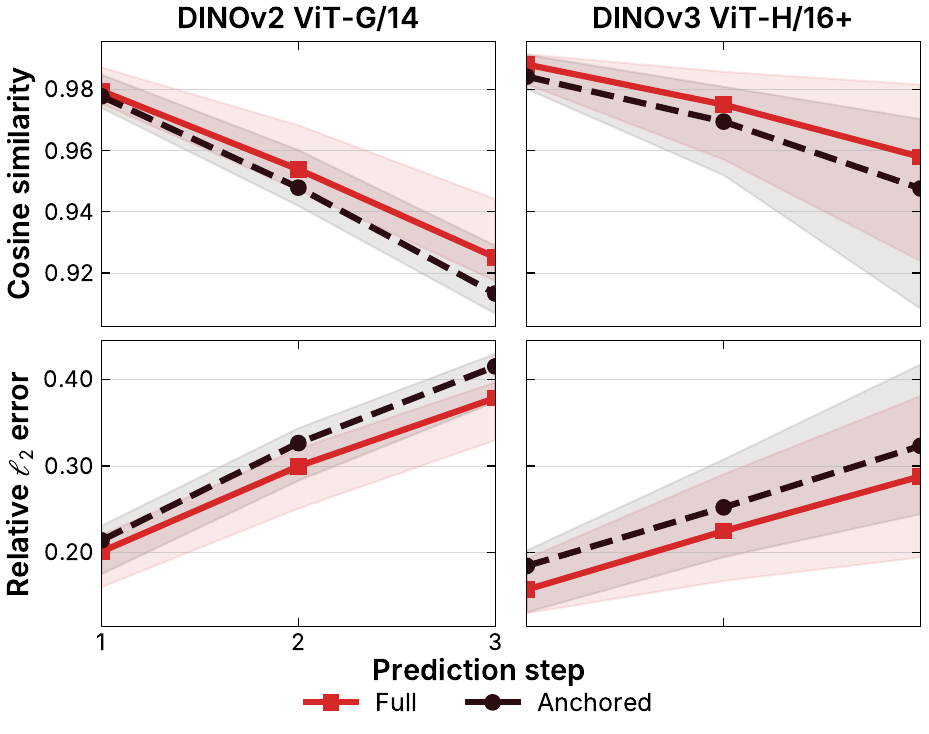}
    \caption{\textbf{\FullDMD vs.\ \anchoredDMD.} Cosine similarity and relative \(\ell_2\) error between predicted and true hidden states at prediction steps \(q \in \{1,2,3\}\), on DINOv2-G/14 and DINOv3-H/16+. Cut starts span \(i \in \{2,\dots,20\}\).}
    \label{fig:fusability_main}
\end{wrapfigure}

\paragraph{Dynamical systems perspective.} Let \(\mX_\ell \in \mathbb{R}^{d \times t \times B}\) denote the output token tensor of block \(\ell \in \{1, \dots, L\}\), where \(d\) is the embedding dimension, \(t\) is the number of tokens, and \(B\) is the number of inputs. Consider the hidden-state trajectory $\{\mX_i, \mX_{i+1}, \ldots, \mX_{i+p}\}$ produced by a contiguous span of $p$ blocks (specifically blocks $i+1, \ldots, i+p$) starting from anchor state $\mX_i$. Treating depth as discrete time, we ask whether this trajectory admits a single, time-invariant linear map $\mK \in \mathbb{R}^{d \times d}$ applied recurrently, $\mX_{i+q} \approx \mK^q \mX_i$ for each step $q \in \{1, \ldots, p\}$. The map acts on the feature axis, so when fitting we flatten the token and batch dimensions. One need only fit $\mK$ to then replace $p$ blocks with a single matrix multiplication, with no retraining. This is the appeal of the \textit{autonomous} hypothesis. Of course, the true ViT is best described as \textit{non-autonomous}, because each block applies a distinct nonlinear map with its own learned parameters. However, we adopt this idealization deliberately, treating the quality of the autonomous linear assumption as the central empirical question of this work. Assuming this perspective leads to several modeling choices, which we discuss below. 
\paragraph{Transformer blocks.} We write a transformer block as
\begin{equation}
\begin{aligned}
\mA_\ell &= \mX_{\ell-1} + \mD^{(1)}_\ell\,\mathrm{MHA}_\ell\!\left(\mathrm{LN}^{(1)}_\ell(\mX_{\ell-1})\right), \\
\mM_\ell &= \mathrm{MLP}_\ell\!\left(\mathrm{LN}^{(2)}_\ell(\mA_\ell)\right), \\
\mX_{\ell} &= \mA_\ell + \mD^{(2)}_\ell\,\mM_\ell,
\end{aligned}
\label{eq:vit_block}
\end{equation}
where \(\mA_\ell\) is the post-attention residual, \(\mM_\ell\) is the MLP output, \(\mathrm{LN}^{(1)}_\ell\) and \(\mathrm{LN}^{(2)}_\ell\) are layer normalizations. \(\mD^{(1)}_\ell, \mD^{(2)}_\ell\) are diagonal LayerScale \citep{touvron_going_2021} matrices, which can be absorbed exactly into the preceding projection weights. We summarize our notation and explain the simple folding procedure in \cref{app:prelim}.

\paragraph{Architecture.}
In fitting $\mK$, we consider two formulations, `\fullDMD' and `\anchoredDMD'. The two approaches make different assumptions about how the linear dynamics are modeled. In the former, given a span of $p$ transformer blocks beginning at index $i$, if we fit $\mK$ using consecutive hidden-state pairs such that
\begin{equation}
\label{eq:full_dmd}
  \mX_{i+p} \approx \mK^p\,\mX_i,
\end{equation}
then we model the evolution of the \textit{full} state \(\mX_i\). Hence, we refer to this formulation as \fullDMD. Alternatively, based on the replacement proposed by \citet{shopkhoev_replaceme_2025} which operates on the MLP residual \(\mD_\ell^{(2)}\mM_\ell\), we also consider \anchoredDMD. Unlike \fullDMD, this formulation fixes \(\mA_i\) as a static \textit{anchor} and fits \(\mK\) to the residual trajectory \(\mX_{i+k} - \mA_i\) for \(k = 0, \ldots, p\), where the initial residual coincides with the kept block's MLP output, \(\mX_i - \mA_i = \mM_i\), yielding
\begin{equation}
\label{eq:anchored_dmd}
  \mX_{i+p} \approx \mA_i + \mK^{p} \mM_i.
\end{equation}
Here, we benefit from folding LayerScale into the preceding MLP weights. Discarding intermediate steps and fitting directly between \(i\) and \(i+p\) would recover \citet{shopkhoev_replaceme_2025}'s method. Although, we note that they do not employ the DMD machinery. Intuitively, $\mA_i$ preserves the input and attention contribution of the kept block, while $\mK$ models how the MLP-driven component evolves across the pruned depth. \FullDMD, on the other hand, directly models the state evolution.

We present a pilot evaluation of both approaches in \cref{fig:fusability_main}. For layers \(i = \{2, 3, \dots, 20\}\), we fit a DMD model with \(p=3\), i.e., it can predict up to \(3\) steps ahead. The figure shows the median cosine similarity and relative \(\ell_2\) error across all images, tokens, and layers, with the shaded regions representing the interquartile range. \FullDMD typically has a slight advantage over \anchoredDMD. However, this advantage does not always carry over to performance in downstream representations, as we demonstrate in \cref{sec:experiments}.

\paragraph{Dynamic mode decomposition.} Agnostic to the architectural choice, we fit \(\mK\) using DMD \citep{schmid_dynamic_2010, kostic_learning_2022} which requires no gradient-based training. We assume \fullDMD for notational simplicity. Consider a network consisting of \(L\) blocks, where we are pruning \(p\) blocks beginning at layer $i$. First, we pass \(B\) calibration images through the entire pretrained network, caching the hidden states $\{\mX_i, \mX_{i+1}, \ldots, \mX_{i+p}\}$ across the target span of transformer blocks. We pool all consecutive snapshot pairs across both images and steps into stacked data matrices $\mZ, \mZ' \in \mathbb{R}^{d \times M}$, where $M = B \cdot p \cdot t$. The matrices \(\mZ\) and \(\mZ'\) are simply time-shifted snapshots. Then, we solve the regression
\begin{equation}
\label{eq:dmd_obj}
  \mK = \arg\min_{\mK \in \mathbb{R}^{d \times d}} \left\| \mZ' - \mK\mZ \right\|_F^2 + \alpha \left\| \mK \right\|_F^2.
\end{equation}
Setting $\alpha = 0$ recovers the least-squares formulation of DMD \citep{schmid_dynamic_2010, tu_dynamic_2014}. Often only discussed in kernel DMD literature, setting the ridge penalty $\alpha > 0$ is an optional approach to regularizing the fitted map \citep{kostic_learning_2022}. We find that performance degrades at strong ridge penalties and there is insignificant benefit at lower penalties. We provide more details on DMD in \cref{app:dmd_explain}. 

\begin{figure}[!tb]
    \centering
    \includegraphics[width=\textwidth]{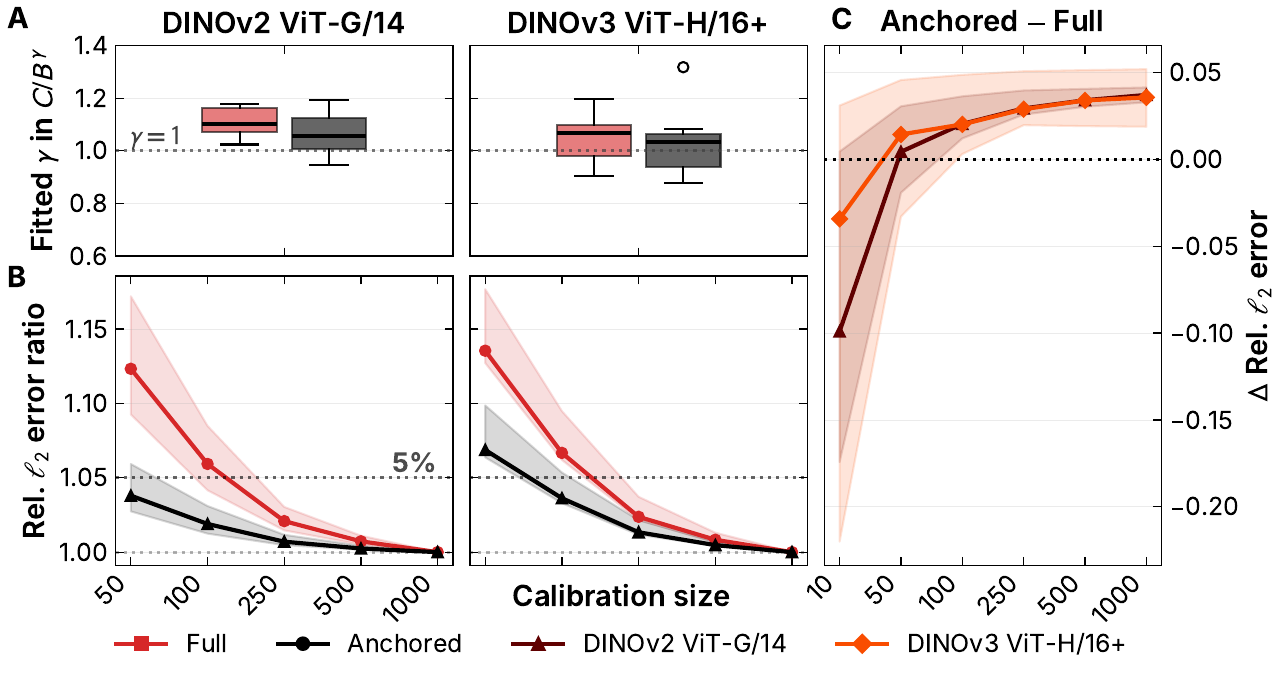}
    \caption{\textbf{Calibration budget and convergence rate.} \textbf{A.}~Fitted exponent \(\gamma\) in the power-law model \(C / B^{\gamma}\) for the relative \(\ell_2\) error, shown as box plots over all configurations (cut starts \(\ell \in \{1,4,7,11,15,18,21,25\}\), prune length fixed at \(3\)); dotted lines mark \(\gamma = 1\). \textbf{B.}~Relative \(\ell_2\) error normalized by the \(B = 1{,}000\) value; the dotted line marks the 5\% threshold. \textbf{C.}~Relative \(\ell_2\) error difference (anchored minus full) as a function of calibration size \(B\).}
    \vspace{-1em}
    \label{fig:calib_gamma}
\end{figure}

\paragraph{Calibration size.} \Cref{eq:dmd_obj} fits a \(d \times d\) operator from \(M = B \cdot p \cdot t\) snapshot pairs. Of the three factors determining \(M\), the token count \(t\) is fixed by the input resolution and patch size and the prune length \(p\) is fixed by the pruning configuration. This leaves the number of calibration images \(B\) as the primary lever controlling the sample size. More concretely, for DINOv3-H/16+ (\(d = 1{,}280\), \(t = 197\)), a single image over a span of \(p = 3\) layers contributes \(M = 3 \times 197 = 591\) snapshot pairs. Thus, for most choices of prune length, modest values of \(B\) place the regression comfortably in the overdetermined regime (\(M \gg d\)). However, an overdetermined system does not guarantee that the calibration set is of adequate size. We fit a power-law model \(C / B^{\gamma}\) to the relative \(\ell_2\) error across all pruning configurations. As shown in \cref{fig:calib_gamma}A, the fitted exponent concentrates tightly around \(\gamma \approx 1\) for both \fullDMD and \anchoredDMD across two models. In practice, both formulations converge to within 5\% of their large-budget baseline by \(B = 250\) images (\cref{fig:calib_gamma}B) and \anchoredDMD is less sensitive to small calibration budgets than \fullDMD (\cref{fig:calib_gamma}C); it converges faster. In \cref{app:calib}, we also demonstrate that the error decays at \(O(d/Bt)\). We conclude that \(p\) does not contribute meaningfully to the effective number of samples. In all further experiments, we use a calibration set of \(1{,}000\) images.

\paragraph{Matrix rank.} One of the main appeals of DMD is a built-in approach to rank truncation of the learned operator. The objective in \cref{eq:dmd_obj} can be modified to specify \(\mathrm{rank}(\mK) \leq r\). DMD literature has seen competing approaches to satisfy this constraint, of which we consider the standard DMD algorithm \citep{schmid_dynamic_2010} and reduced rank regression (RRR) \citep{kostic_learning_2022}. The algorithm proposed by \citet{schmid_dynamic_2010} is equivalent to principal component regression (PCR) \citep{kostic_learning_2022, brunton_modern_2022}, which we reproduce in \cref{app:dmd_explain}. Examining this connection clarifies that standard DMD can be aptly summarized as regressing in the coordinates defined by \(\mU\), where \(\mZ = \mU \mSigma \mV^\top\). Thus, finding a low-rank \(\mK\) involves truncating the left singular vectors of the input matrix. On the other hand, RRR involves truncating a different object, namely \(\mZ'\mZ^{\top} \left( \mZ \mZ^{\top} \right)^{-1/2}\). Broadly speaking, the latter object contains information from both the input \(\mZ\) and output \(\mZ'\), which may explain why \citet{kostic_learning_2022} are able to demonstrate RRR's improvement over PCR. Our experiments rely on the \texttt{kooplearn} library \citep{turri_kooplearn_2026}, which implements both PCR and RRR, allowing us to compare both algorithms for rank selection. We provide a head-to-head in \cref{app:rrr}.
{
\begin{figure}[!b]
\centering
{\footnotesize
\captionof{table}{Cosine similarity\,$\uparrow$\;/\;relative $\ell_2$ error\,$\downarrow$ of the predicted hidden state, reported as the median over all cut-start positions.}
\label{tab:headline}
\begin{tabular}{l c c c c c c}
\toprule
Method & $p=1$ & $p=3$ & $p=4$ & $p=5$ & $p=7$ & $p=10$ \\
\midrule
\multicolumn{7}{l}{\textbf{DINOv3-H/16+}} \\
Full DMD & 0.98\,/\,0.17 & 0.95\,/\,0.31 & 0.91\,/\,0.42 & 0.87\,/\,0.50 & 0.82\,/\,0.61 & 0.68\,/\,0.76 \\
Anchored DMD & 0.98\,/\,0.18 & 0.94\,/\,0.33 & 0.92\,/\,0.40 & 0.88\,/\,0.47 & 0.78\,/\,0.63 & 0.66\,/\,0.76 \\
ReplaceMe & 0.98\,/\,0.18 & 0.95\,/\,0.31 & 0.93\,/\,0.37 & 0.90\,/\,0.43 & 0.83\,/\,0.56 & 0.74\,/\,0.67 \\
Identity & 0.97\,/\,0.31 & 0.87\,/\,0.56 & 0.84\,/\,0.63 & 0.79\,/\,0.68 & 0.69\,/\,0.76 & 0.56\,/\,0.85 \\
\addlinespace
\multicolumn{7}{l}{\textbf{DINOv2-G/14}} \\
Full DMD & 0.97\,/\,0.24 & 0.91\,/\,0.41 & 0.87\,/\,0.48 & 0.83\,/\,0.55 & 0.75\,/\,0.67 & 0.61\,/\,0.82 \\
Anchored DMD & 0.97\,/\,0.26 & 0.90\,/\,0.43 & 0.86\,/\,0.51 & 0.82\,/\,0.59 & 0.72\,/\,0.70 & 0.59\,/\,0.82 \\
ReplaceMe & 0.98\,/\,0.22 & 0.91\,/\,0.41 & 0.87\,/\,0.48 & 0.84\,/\,0.55 & 0.76\,/\,0.65 & 0.65\,/\,0.76 \\
Identity & 0.97\,/\,0.25 & 0.89\,/\,0.49 & 0.83\,/\,0.58 & 0.78\,/\,0.65 & 0.66\,/\,0.77 & 0.52\,/\,0.87 \\
\bottomrule
\end{tabular}}

\smallskip

\includegraphics[width=0.9\textwidth]{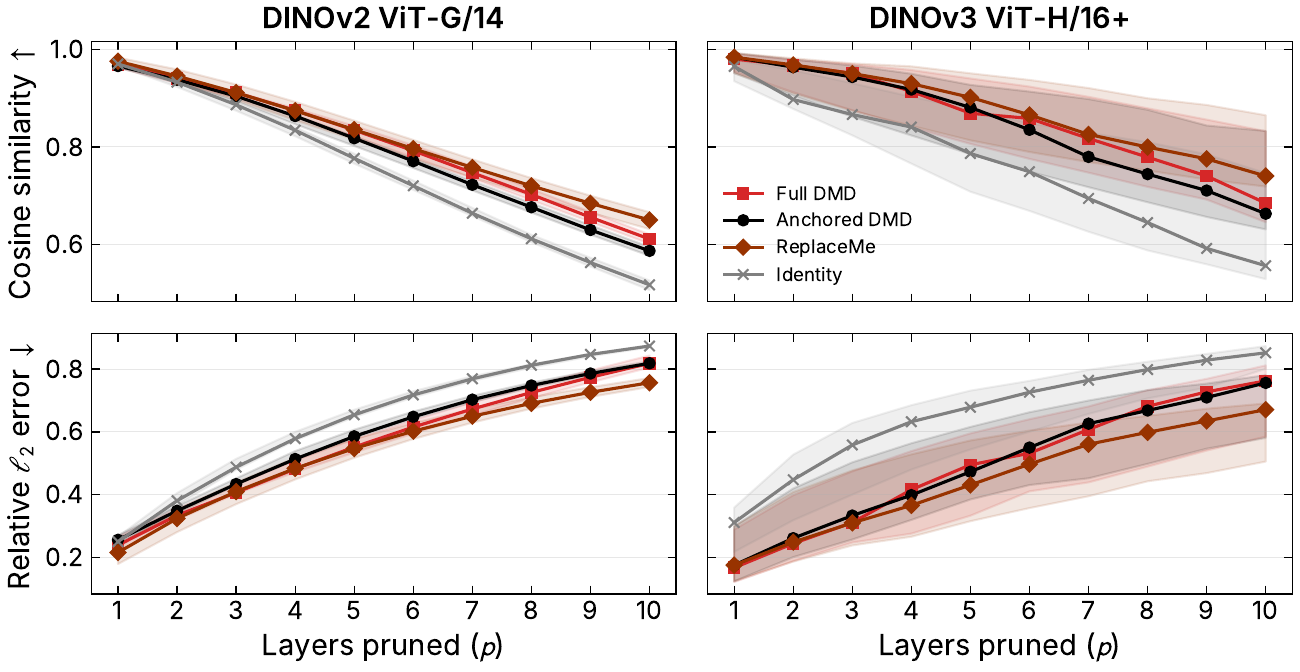}
\caption{\textbf{Reconstruction quality across prune lengths.} Per-position breakdown of the metrics in Table~\ref{tab:headline}, across all prune lengths $p$.}
\label{fig:headline}
\end{figure}
}

\FloatBarrier

\section{Experiments}
\label{sec:experiments}

\paragraph{Cost of the autonomous hypothesis.} We study the quality of the linear and autonomous hypothesis on four pretrained DINO ViTs, fitting linear maps with \fullDMD, \anchoredDMD, and ReplaceMe \citep{shopkhoev_replaceme_2025}, plus an identity baseline. At each valid cut start \(i\), we fit \(10\) operators, each optimized to predict \(\mX_{i+p}\) for \(p \in \{1, \dots, 10\}\). For DMD, a single fitted \(\mK\) yields predictions for all spans \(q \leq p\) via \(\mK^q\). \Cref{tab:headline} and \cref{fig:headline} report median cosine similarity and relative \(\ell_2\) error across all cut-starts, for selected \(p\) in the table and all \(p\) in the figure, along with IQR bands; results for DINOv2/v3-L variants are in \cref{app:sweep_details}. At \(p{=}1\), all three learned methods achieve cosine similarity above \(0.97\) on DINOv3-H/16+, in line with \citet{razzhigaev_your_2024}. As the prune length grows, the methods separate. As expected, ReplaceMe generally attains the highest reconstruction quality across both metrics, as it avoids midpoint constraints. The gap between ReplaceMe and the DMD methods therefore quantifies the cost of the autonomous hypothesis, i.e., what is lost by requiring the dynamics to factorize as a single operator \(\mK\) applied repeatedly. At short prune lengths (\(p \leq 4\)), this cost is modest. \FullDMD reaches cosine within \(0.02\) of ReplaceMe on DINOv3-H/16+, indicating that the autonomous constraint is approximately satisfied in this regime. As \(p\) grows, the gap widens, suggesting depth-dependent dynamics that a single time-invariant operator cannot fully capture. Friedman tests (see \cref{app:sweep_details}), as recommended in \citet{demsar_statistical_2006}, confirm a statistically significant ordering and performance difference across all methods. \FullDMD beats \anchoredDMD on cosine in all four models; on relative \(\ell_2\) error the DMD variant ordering is model-dependent.

\begin{figure}[!h]
    \centering
    \includegraphics[width=0.9\textwidth]{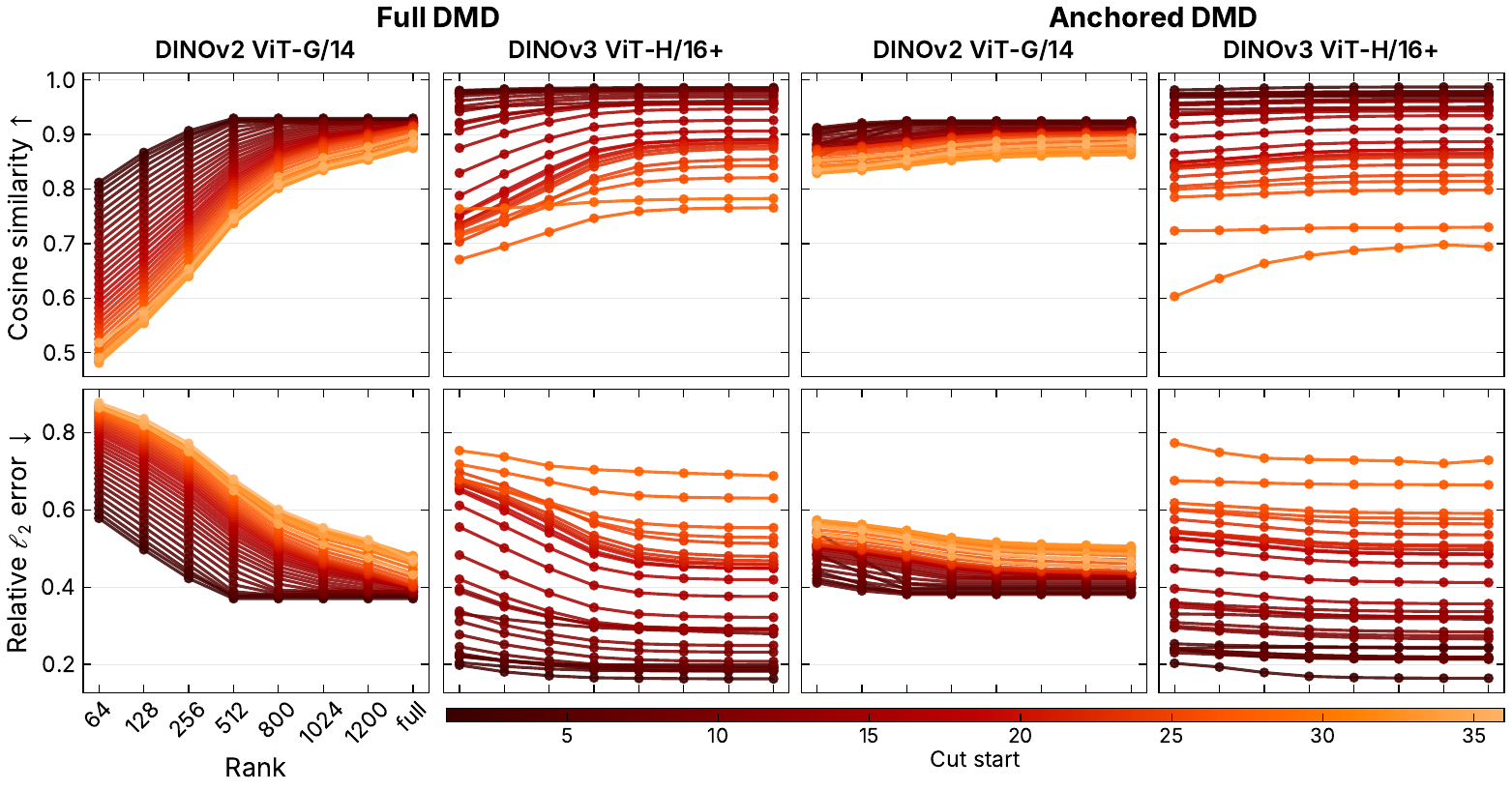}
    \caption{\textbf{Reconstruction quality vs operator rank.} Cosine similarity (top) and relative \(\ell_2\) error (bottom) of the predicted hidden state, as a function of operator rank for \fullDMD (left two columns) and \anchoredDMD (right two columns) on DINOv2-G/14 and DINOv3-H/16+. Each line is a cut start, colored by depth. Operators fit with RRR; PCR yields equivalent results within \(\pm 0.01\) (\cref{fig:pcr_vs_rrr}).}
    \vspace{-1.5em}
    \label{fig:rank}
\end{figure}
\paragraph{Rank selection.} \Cref{fig:rank} shows that fitted operators can admit substantial rank reduction, especially for earlier layers. On DINOv3-H/16+, early cut starts (\(i \leq 7\)) match their full-rank cosine similarity already at rank 64 under both formulations, suggesting the one-step dynamics concentrate in a subspace of dimension \(\ll d\). This compressibility decays monotonically with cut depth, with late cut starts requiring higher rank to saturate. The pattern is consistent across both models, highlighting that early transformer blocks are easier to linearize than later ones. DINOv2-G/14 cosine similarity plateaus near \(0.92\), below the \(0.97+\) reached on DINOv3-H/16+, indicating DINOv2-G/14 is less amenable to linearization overall.
\begin{figure}[!t]
    \centering
    \includegraphics[width=\textwidth]{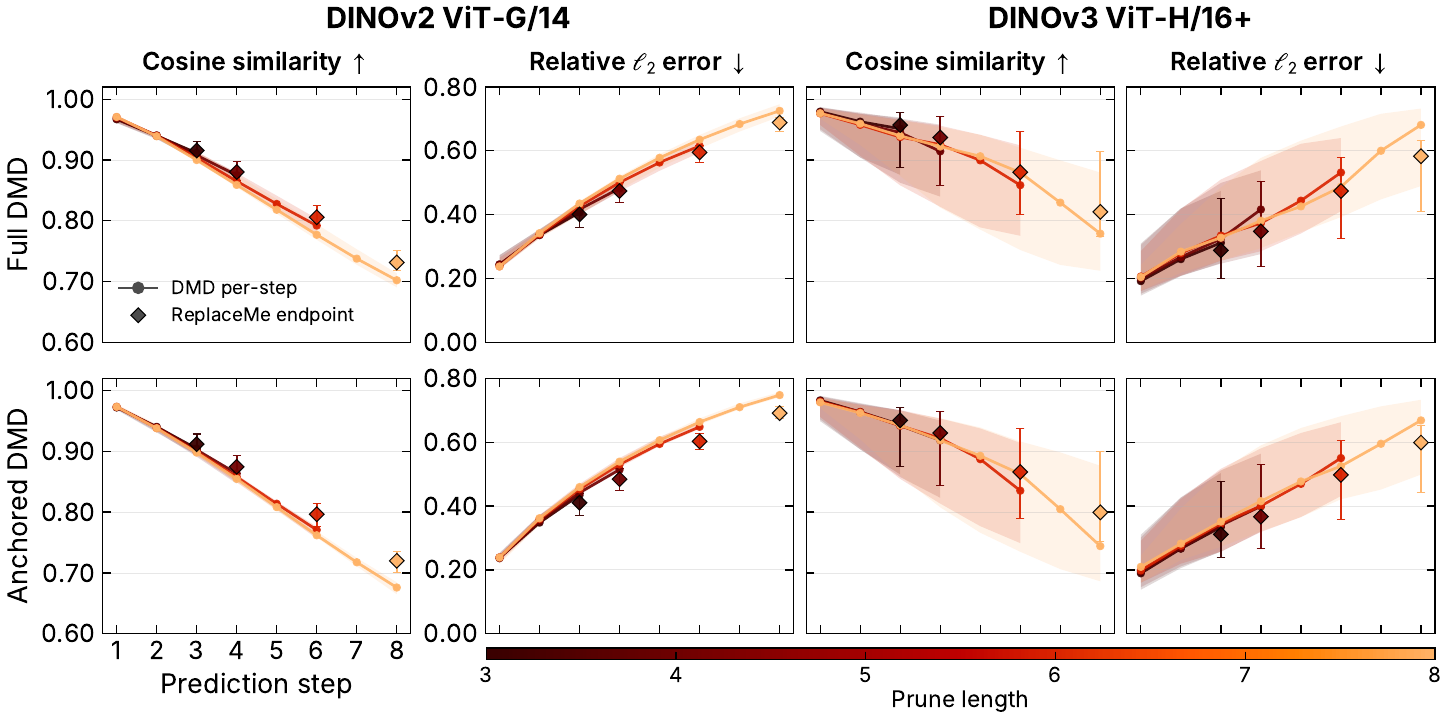}
    \caption{\textbf{Intermediate-step prediction quality.} Cosine similarity between \(\mK^q \mX_i\) and the true hidden state \(\mX_{i+q}\) at each intermediate step \(q\), on DINOv2-G/14 (left) and DINOv3-H/16+ (right). ReplaceMe is shown as reference at \(q = p\) only.}
    \vspace{-1.5em}
    \label{fig:rollout}
\end{figure}
\begin{figure}[!b]
    \centering
    \includegraphics[width=\textwidth]{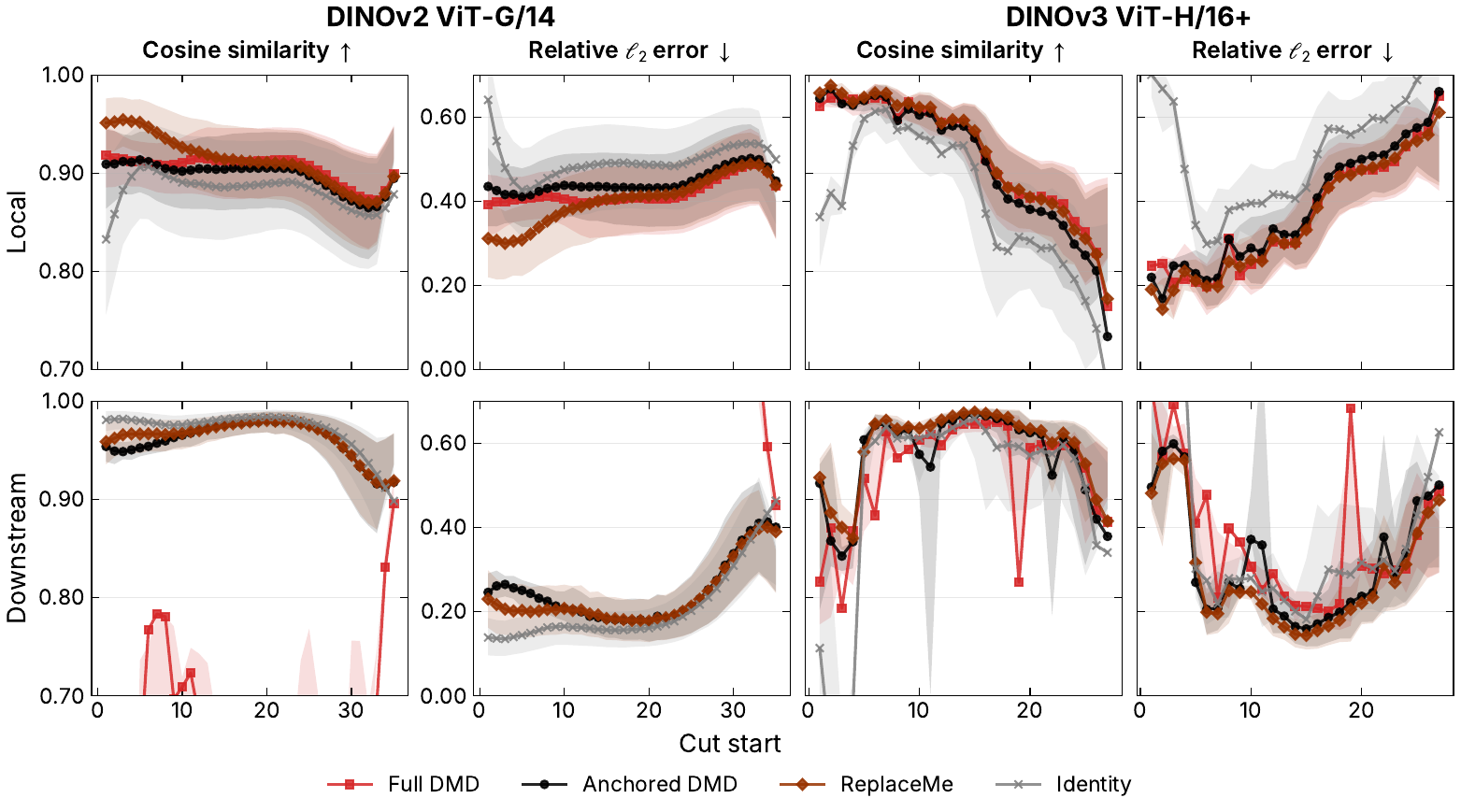}
    \caption{\textbf{Local vs downstream error of linear operators.} Cosine similarity (left within each model) and relative \(\ell_2\) error (right within each model) for \fullDMD, \anchoredDMD, ReplaceMe, and an identity baseline on DINOv2-G/14 (left two columns) and DINOv3-H/16+ (right two columns) at prune length \(p=3\). The top row (\emph{Local}) shows performance at the operator output \(\mX_{i+p}\); the bottom row (\emph{Downstream}) shows performance at the final hidden state \(\mX_L\).}
    \label{fig:robustness}
\end{figure}
\paragraph{Intermediate activations.} The metrics in \cref{tab:headline} measure DMD's prediction quality only at the cut-endpoint \(\mX_{i+p}\), but a fitted \(\mK\) also produces predictions \(\mK^q \mX_i\) at every intermediate step \(q \in \{1,\ldots,p\}\). \Cref{fig:rollout} compares \(\mK^q \mX_i\) against the true \(\mX_{i+q}\). Quality decays monotonically with \(q\), mirroring the cut-endpoint pattern in \cref{fig:headline}. We also plot the direct endpoint map from ReplaceMe as reference.

\paragraph{Downstream representations.} A natural follow-up question is whether the linear maps preserve their reconstruction quality through the remaining transformer blocks. \Cref{fig:robustness} compares each method's error at two locations, the operator output \(\mX_{i+p}\) (top row, \emph{Local}) and the final hidden state \(\mX_L\) after running the remaining blocks (bottom row, \emph{Downstream}). The top row reflects results from \cref{fig:headline}, with all three learned methods better than the identity baseline at the cut on both metrics. The bottom row shows that \FullDMD fails downstream, sometimes catastrophically, with cosine similarity collapsing to \(\sim 0.78\) over a wide band of cut starts on DINOv2-G/14 and dropping sharply at several cut starts on DINOv3-H/16+. Surprisingly, the identity baseline is competitive with, or better than, both \anchoredDMD and ReplaceMe at the final hidden state, indicating that local reconstruction quality does not predict downstream behavior in DINO models.

\begin{wrapfigure}{r}{0.5\textwidth}
    \centering
    \includegraphics[width=0.5\textwidth]{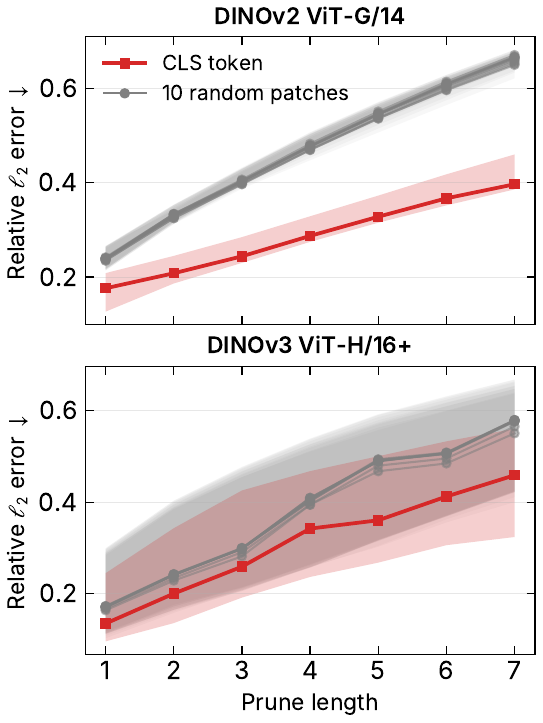}
    \caption{\textbf{CLS privilege.} Relative \(\ell_2\) error of \fullDMD predictions per token type across \(p\), using the operators fit in the headline sweep (no retraining).}
    \label{fig:cls_privilege}
\end{wrapfigure}

\paragraph{CLS privilege.} In line with ReplaceMe, the DMD map fits on all tokens jointly. Here we consider whether those same operators predict different token types equally well. \Cref{fig:cls_privilege} shows that, in both models, the CLS token is consistently more linearly approximable than the patch tokens, which cluster together above the CLS line in both models. While we do not fit per-token operators, we would expect this gap to further increase.

\section{Discussion}

\paragraph{Limitations.}
We focus exclusively on the DINO family of ViTs; whether the autonomous linear hypothesis behaves the same way under other transformers is not investigated here. LLMs are more involved due to the autoregressive nature of language model training; unlike ViTs, tokens have a sequential nature. More broadly, our study is an empirical characterization of when the autonomous linear assumption holds across several design choices (cut depth, span length, calibration, rank, and token type). We do not provide a directly actionable approach to reduce parameter count, nor a principled metric for jointly selecting the cut start and prune length.

\paragraph{Healing.} \citet{shopkhoev_replaceme_2025} report that `healing', i.e., fine-tuning after pruning blocks, can recover downstream performance. We expect this  might close the gap exposed in \cref{fig:robustness}, where local reconstruction quality fails to predict downstream behavior; a few gradient steps after substitution could realign the \(\mK\) output to the regime expected by subsequent blocks. However, that incurs a non-trivial training cost.

\paragraph{Linear dynamics in a lifted space.} Our \(\mK\) is fit in closed form via DMD on the raw hidden states. Enclosing this linear operator within a learned observable space (à la Koopman autoencoders), would lift the dynamics into a richer, high dimensional representation and could capture the depth-dependent structure that fitting on raw states would not.

\paragraph{Transformer block alignment.} \citet{aubry_transformer_2025a}'s idealized linearization \(\mX_\ell = (\mI + \mJ)^\ell \mX_0\) (\cref{eq:aubry-recurrent}) is precisely the form of operator our DMD fit produces. \citet{aubry_transformer_2025a} derive this expression as the limiting behavior implied by residual Jacobian alignment. Our fitted \(\mK\) is the empirical version of this generator, recovered directly from snapshot data without assuming coupling a priori. The match between \fullDMD and the unconstrained ReplaceMe map at short prune lengths supports the autonomous-coupling picture in this regime; the widening gap at longer \(p\) shows that coupling assumption is only approximate, with the small per-step misalignment compounding under repeated application of \(\mK\). Our findings further suggest the coupling is non-uniform along both axes \citet{aubry_transformer_2025a} identify. Across depth, early layers admit lower-rank operators than late layers (\cref{fig:rank}), consistent with stronger alignment near the input. Across tokens, the CLS token is more linearly approximable than patches (\cref{fig:cls_privilege}), suggesting CLS sits in a more strongly coupled subspace.

\paragraph{Block Recurrence} Our work began with the observations from \citet{jacobs_block_2026}, where they replace a span of blocks with a single transformer block, which requires a significant training effort to learn the block's parameters. Future work might explore whether such block-recurrent models are more or less amenable to DMD fits.

\newpage
\section*{Acknowledgments}
This work was supported by the NYUAD Center for Interacting Urban Networks (CITIES), funded by Tamkeen under the NYUAD Research Institute Award CG001. The views expressed in this article are those of the authors and do not reflect the opinions of CITIES or their funding agencies.
\bibliographystyle{apalike}
\bibliography{refs}    

\newpage
\appendix
\crefalias{section}{appendix}
\section{Preliminaries}
\label{app:prelim}

To orient the discussion, \cref{tab:notation} summarizes the notation used throughout our work. For completeness, \cref{app:transformer,app:fold-ls} present a brief summary of the Vision Transformer (ViT) architecture.

\begin{table}[h]
\centering
\renewcommand{\arraystretch}{1.2}
\begin{tabular}{ll}
\toprule
Symbol & Meaning \\
\midrule
\(L\) & Total number of transformer blocks. \\
\(\ell \in \{1,\dots,L\}\) & Transformer block index. \\
\(t\) & Total number of tokens (cls, register, and patch). \\
\(d\) & Token embedding dimension. \\
\(d_{\mathrm{ff}}\) & MLP hidden dimension. \\
\(\mX_\ell \in \mathbb{R}^{d \times t \times B}\) & Output activations of block \(\ell\). \\
\(\mA_\ell \in \mathbb{R}^{d \times t \times B}\) & Post-attention residual activations in block \(\ell\), i.e.\ \(\mX_{\ell-1} + \mathrm{MHA}_\ell(\cdot)\). \\
\(\mM_\ell \in \mathbb{R}^{d \times t \times B}\) & MLP activations in block \(\ell\). \\
\(\mathrm{LN}^{(1)}_\ell,\mathrm{LN}^{(2)}_\ell\) & Pre-attention and pre-MLP layer normalization layers. \\
\(\mathrm{MHA}_\ell\) & Attention map in block \(\ell\). \\
\(\mathrm{MLP}_\ell\) & MLP map in block \(\ell\). \\
\(\mD^{(1)}_\ell, \mD^{(2)}_\ell \in \mathbb{R}^{d \times d}\) & LayerScale diagonal matrices for attention and MLP. \\
\(\mI\) & Identity matrix; size inferred from context. \\
\(\mathbbm{1}_d\) & \(d\)-sized vector of all \(1\)s. \\
\(p\) & Prune length. \\
\bottomrule
\end{tabular}%
\caption{Summary of notation.}
\label{tab:notation}
\end{table}
\subsection{Transformer Blocks}
\label{app:transformer}
Let \(\mX_\ell \in \mathbb{R}^{d \times t \times B}\) denote the output activations of any block \(\ell \in \{1, \dots, L\}\) over \(B\) inputs. Reproducing \cref{eq:vit_block}, we write the operations of a ViT block as
\begin{equation*}
\begin{aligned}
\mA_\ell &= \mX_{\ell-1} + \mD^{(1)}_\ell\,\mathrm{MHA}_\ell\!\left(\mathrm{LN}^{(1)}_\ell(\mX_{\ell-1})\right), \\
\mM_\ell &= \mathrm{MLP}_\ell\!\left(\mathrm{LN}^{(2)}_\ell(\mA_\ell)\right), \\
\mX_{\ell} &= \mA_\ell + \mD^{(2)}_\ell\,\mM_\ell.
\end{aligned}
\end{equation*}

\subsection{Folding LayerScale}
\label{app:fold-ls}

ViTs typically include LayerScale \citep{touvron_going_2021}, a learnable diagonal matrix applied after each residual component (attention or MLP). LayerScale can be absorbed exactly into the preceding component. Using the MLP as an example, let \(\mH_\ell \in \mathbb{R}^{d_{\mathrm{ff}} \times t \times B}\) denote the hidden state after the nonlinearity, with output projection weight \(\mW_\ell^{\mathrm{out}} \in \mathbb{R}^{d \times d_{\mathrm{ff}}}\) and bias \(\vb_\ell^{\mathrm{out}} \in \mathbb{R}^{d}\). Since \(\mD_\ell^{(2)}\) is diagonal, left-multiplication distributes through the affine map,
\begin{equation}
\mD_\ell^{(2)}\!\left(\mW_\ell^{\mathrm{out}}\mH_\ell + \vb_\ell^{\mathrm{out}}\right)
=
\underbrace{\mD_\ell^{(2)}\mW_\ell^{\mathrm{out}}}_{\widetilde{\mW}_\ell^{\mathrm{out}}}\mH_\ell
+
\underbrace{\mD_\ell^{(2)}\vb_\ell^{\mathrm{out}}}_{\widetilde{\vb}_\ell^{\mathrm{out}}}.
\end{equation}
The same applies to the attention branch. After folding, the block takes the standard pre-norm residual form,
\begin{equation}
\begin{aligned}
\mA_\ell &= \mX_{\ell-1} + \widetilde{\mathrm{MHA}}_\ell\!\left(\mathrm{LN}^{(1)}_\ell(\mX_{\ell-1})\right), \\
\mM_\ell &= \widetilde{\mathrm{MLP}}_\ell\!\left(\mathrm{LN}^{(2)}_\ell(\mA_\ell)\right), \\
\mX_{\ell} &= \mA_\ell + \mM_\ell.
\end{aligned}
\label{eq:block-folded}
\end{equation}
This reparameterization is exact. In all experiments, we fold LayerScale to simplify our analysis.
\section{Background}
\label{app:background}

\paragraph{ReplaceMe.} \citet{shopkhoev_replaceme_2025} propose \textit{ReplaceMe}, a method for the `structural depth pruning' of transformers. Their method identifies a contiguous set of transformer blocks and replaces them with a linear map. For a fixed prune length \(p\), ReplaceMe selects a cut index \(i_\star\) by measuring the discrepancy between block outputs before and after a candidate span (contiguous blocks). Using our notation, the selected cut is
\begin{equation}
i^\star = \arg\min_\ell h\!\left(\mX_{\ell}, \mX_{\ell+p}\right),
\label{eq:rm-cut}
\end{equation}
where \(h(\cdot,\cdot)\) is a measure of distance between hidden states, averaged over a calibration set. Once the cut index \(i^\star\) has been chosen, ReplaceMe removes blocks \(\{i^\star+1,\dots,i^\star+p\}\) and estimates a linear map \(\mT \in \mathbb{R}^{d \times d}\) acting on the MLP branch of block \(i^\star\) by solving
\begin{equation}
\mT^\star
\in
\arg\min_{\mT}
h\!\left(\mA_{i^\star} + \mT \mM_{i^\star},\mX_{i^\star+p}\right).
\label{eq:rm-obj}
\end{equation}
The compensated block output is then
\begin{equation}
\widehat{\mX}_{\ell+p} = \mA_\ell + \mT^\star \mM_\ell.
\label{eq:rm-deploy}
\end{equation}
Because \(\mT^\star\) acts linearly on the channel dimension, it can be fused exactly into the MLP output projection, in a manner similar to that described in \cref{app:fold-ls}. To elaborate,
\begin{equation}
\mM_\ell = \widetilde{\mW}_\ell^{\mathrm{out}} \mH_\ell + \widetilde{\vb}_\ell^{\mathrm{out}},
\label{eq:rm-mlp-out}
\end{equation}
where \(\mH_\ell \in \mathbb{R}^{d_{\mathrm{ff}} \times t \times B}\) is the post-nonlinearity MLP hidden state. Then
\begin{equation}
\begin{aligned}
\mT^\star \mM_\ell
&=
\mT^\star\!\left(\widetilde{\mW}_\ell^{\mathrm{out}} \mH_\ell + \widetilde{\vb}_\ell^{\mathrm{out}}\right) \\
&=
\left(\mT^\star \widetilde{\mW}_\ell^{\mathrm{out}}\right) \mH_\ell
+
\mT^\star \widetilde{\vb}_\ell^{\mathrm{out}} .
\end{aligned}
\label{eq:rm-fuse-expand}
\end{equation}
Therefore the transformed branch is implemented exactly by
\begin{equation}
\widetilde{\mW}_\ell^{\mathrm{out,RM}}
=
\mT^\star \widetilde{\mW}_\ell^{\mathrm{out}},
\qquad
\widetilde{\vb}_\ell^{\mathrm{out,RM}}
=
\mT^\star \widetilde{\vb}_\ell^{\mathrm{out}}.
\label{eq:rm-fuse}
\end{equation}
Thus, \(\mT^\star\) approximates the action of a pruned span and is fused into the existing block.

\paragraph{Block-Recurrent Hypothesis.} \citet{jacobs_block_2026} propose the \emph{Block-Recurrent Hypothesis (BRH)}, which posits that the intermediate computations of a depth-\(L\) ViT can be reproduced by repeatedly applying a small set of parameter-tied blocks. Formally, a model satisfies the \(\varepsilon\)-BRH if, for every \(\ell\), there exist blocks \(B_1,\dots,B_k\) with \(k \ll \ell\) and positive integers \(n_1,\dots,n_k\) with \(\sum_{j=1}^k n_j = \ell\) such that
\begin{equation}
\mathbb{E}_{x \sim \mathcal P}
\left[
\left\|
\mX_\ell -
\left(B_k^{(n_k)} \circ \cdots \circ B_1^{(n_1)}\right)(x)
\right\|_F
\right]
\le \varepsilon,
\label{eq:brh}
\end{equation}
where \(B_j^{(n_j)}\) denotes \(n_j\) consecutive applications of the parameter-tied block \(B_j\), and \(\mathcal P\) is a distribution over natural images. Requiring \cref{eq:brh} at every \(\ell\) (not only \(\ell=L\)) forces the surrogate to track the trajectory of hidden states \(\mX_1,\dots,\mX_L\) layer by layer. \citet{jacobs_block_2026} operationalize their hypothesis by training weight-tied block-recurrent surrogates, with phase boundaries determined by a max-cut segmentation of the layer-layer similarity matrix. Let \(\widetilde{\mX}_\ell\) denote the surrogate hidden state at depth \(\ell\), produced autoregressively by the tied blocks. The training objective enforces trajectory fidelity up to a horizon \(h \le L\),
\begin{equation}
\mathcal{L}^{\mathrm{AR}}_h
=
\mathbb{E}_{x}
\left[
\sum_{\ell=1}^{h}
\left\|
\widetilde{\mX}_\ell - \mX_\ell
\right\|_F
\right],
\label{eq:raptor-loss}
\end{equation}
so that the surrogate matches the reference activations along the first \(h\) blocks.

\paragraph{Residual Alignment.}
\citet{li_residual_2023} linearize each residual block by its residual Jacobian
\begin{equation}
\mJ_\ell = \frac{\partial f_\ell}{\partial \mX_{\ell-1}},
\qquad
\mJ_\ell = \mU_\ell\,\mathbf{\Sigma}_\ell\,\mV_\ell^\top,
\label{eq:ra-jacobian}
\end{equation}
where \(f_\ell\) is the non-skip branch of block \(\ell\), so that the residual update reads \(\mX_\ell = (\mI + \mJ_\ell)\,\mX_{\ell-1}\). They identify \emph{Residual Alignment (RA)}, an observation that the top singular vectors of \(\mJ_\ell\) align across depth in well-generalizing ResNets, vanishing once skip connections are removed. Alignment is obtained when \(\mU_j^{\top}\mJ_i\mV_j\) is  diagonal, where \(\mU_i, \mV_j\) are the left and right singular vectors of the \(i\)-th residual Jacobian. \citet{aubry_transformer_2025a} extend this analysis to pretrained LLMs. They confirm singular-vector alignment of block Jacobians across depth and uncover an additional axis of \emph{coupling}, namely that top singular vectors also align across tokens within each block. Most relevant to our work, their Appendix A.5 lays out a \emph{dynamical motivation} for alignment. Each transformer block can be written as a discrete-time difference equation \(\mX_{\ell+1} - \mX_\ell = \mJ_\ell\,\mX_\ell\), so the unrolled state at depth \(L\) factors as
\begin{equation}
\mX_L = \prod_{\ell=1}^{L} (\mI + \mJ_\ell)\, \mX_0,
\label{eq:aubry-product}
\end{equation}
which without further assumptions is the path-decomposition view of \citet{veit_residual_2016}. To unpack the role of coupling, suppose all \(\mJ_\ell\) share singular vectors, \(\mJ_\ell = \mU \mathbf{\Sigma}_\ell \mU^\top\). Then the product collapses to \(\mX_L = \mU\bigl(\prod_\ell (\mI + \mathbf{\Sigma}_\ell)\bigr) \mU^\top \mX_0\), and if additionally the singular values are consistent across depth (\(\mathbf{\Sigma}_\ell = \mathbf{\Sigma}\)), the system reduces to a single recurrent operator,
\begin{equation}
\mX_\ell = (\mI + \mJ)^\ell\, \mX_0.
\label{eq:aubry-recurrent}
\end{equation}
\citet{aubry_transformer_2025a} use the resulting exponential spacing of intermediate hidden states as an indirect quantitative test for coupling, and report that the strength of this coupling correlates with downstream model performance more strongly than parameter count, depth, or embedding dimension alone.

The recurrent linearization in \cref{eq:aubry-recurrent} is precisely the form of operator we fit. Where \citet{aubry_transformer_2025a} derive \((\mI + \mJ)^\ell\) as the limiting behavior implied by spectral coupling on per-block Jacobians, we fit a single operator \(\mK\) directly from snapshot data using DMD, with \(\mX_{i+p} \approx \mK^p \mX_i\) over a chosen span. Our \(\mK\) approximates the time-invariant generator implied by their constant-coupling limit, but obtained from data without first assuming Jacobian coupling. Additionally, DMD allows for a built-in approach to reduce the dimensionality of the fitted map.
\section{Experimental Procedure and Metrics}
\label{app:metrics}

\subsection{Experimental procedure}
\label{app:procedure}

\paragraph{Data.}
All experiments draw images at random from ImageNet-1K. Unless otherwise stated, each operator is fit on \(1{,}000\) calibration images sampled from the training split and evaluated on \(2{,}000\) held-out images sampled from the validation split. Each image is resized to \(256\) pixels, center-cropped to \(224\) pixels, and normalized using per-channel mean \([0.485, 0.456, 0.406]\) and standard deviation \([0.229, 0.224, 0.225]\). For a layer \(\ell\), we compare predicted and ground-truth activations
\[
\hat{\mX}_\ell, \mX_\ell \in \mathbb{R}^{d \times t \times B},
\]
where \(d\) is the feature dimension, \(t\) is the number of tokens, and \(B\) is the batch size.

\paragraph{Operator fitting defaults.}
Unless otherwise stated, all linear operators are fit at full rank using ridge-regularized regression with \(\alpha = 10^{-5}\). Where rank truncation is applied, we use reduced-rank regression (RRR) by default; principal component regression (PCR) is reported in \cref{app:rrr} as a comparison. The DMD machinery is implemented via the \texttt{kooplearn} library \citep{turri_kooplearn_2026}.

\paragraph{Plotting conventions.}
A single metric value at cut start \(i\) is itself an average over the validation set, taken token-wise and image-wise per the definitions below. When we plot across cut starts, solid lines show the median of these per-cut-start values, and shaded bands give the interquartile range (25--75\%) of the same per-cut-start values. The aggregating dimension is typically cut start; deviations are noted in the relevant caption.

\paragraph{Hardware and software.}
All experiments are run on a single NVIDIA RTX 6000 Ada Generation GPU (48~GB), using Python 3.13.9, PyTorch 2.10.0 with CUDA 12.8, and \texttt{kooplearn} 2.0.2.

\subsection{Metrics}

We report several complementary metrics.

\paragraph{Cosine similarity.}
We measure the average token-wise cosine similarity:
\begin{equation}
\mathrm{CosSim}(\hat{\mX},\mX)
=
\frac{1}{Bt}\sum_{b=1}^{B}\sum_{\tau=1}^{t}
\frac{\langle \hat{\vx}_{b,\tau}, \vx_{b,\tau}\rangle}
{\|\hat{\vx}_{b,\tau}\|_2\,\|\vx_{b,\tau}\|_2}.
\end{equation}

\paragraph{Relative \(\ell_2\) error.}
We also measure the average token-wise relative reconstruction error:
\begin{equation}
\label{eq:rel_l2}
\ell_2(\hat{\mX},\mX)
=
\frac{1}{Bt}\sum_{b=1}^{B}\sum_{\tau=1}^{t}
\frac{\|\hat{\vx}_{b,\tau}-\vx_{b,\tau}\|_2}{\|\vx_{b,\tau}\|_2}.
\end{equation}

\paragraph{\(R^2\) score.}
Following \citet{jacobs_block_2026}, we compute the following. Let
\[
\mP = \mI - \frac{1}{d}\mathbbm{1}_d \mathbbm{1}_d^\top .
\]
Then
\begin{equation}
R^2_{\mathrm{BRH}}(\hat{\mX},\mX)
=
\frac{1}{Bt}\sum_{b=1}^{B}\sum_{\tau=1}^{t}
\frac{\langle \mP \hat{\vx}_{b,\tau}, \mP \vx_{b,\tau}\rangle^2}
{\|\mP \hat{\vx}_{b,\tau}\|_2^2\,\|\mP \vx_{b,\tau}\|_2^2}.
\end{equation}

\paragraph{Norm ratio.}
To assess whether the predicted activations preserve magnitude, we report the average norm ratio:
\begin{equation}
\mathrm{NormRatio}(\hat{\mX},\mX)
=
\frac{1}{Bt}\sum_{b=1}^{B}\sum_{\tau=1}^{t}
\frac{\|\hat{\vx}_{b,\tau}\|_2}{\|\vx_{b,\tau}\|_2}.
\end{equation}

\section{Dynamic Mode Decomposition (DMD)}
\label{app:dmd_explain}
\subsection{Introduction}
\begin{algorithm}[bhtp]
\caption{Standard DMD \citep{schmid_dynamic_2010}; presentation adapted from \citet{tu_dynamic_2014}.}
\label{alg:standard_dmd}
\begin{algorithmic}[1]

\STATE Form the data matrices
\[
\mZ = \begin{bmatrix}
\vx_0 & \vx_1 & \cdots & \vx_{m-1}
\end{bmatrix},
\qquad
\mZ' = \begin{bmatrix}
\vx_1 & \vx_2 & \cdots & \vx_m
\end{bmatrix}.
\]

\STATE Compute the reduced singular value decomposition of $\mZ$:
\[
\mZ = \mU_r \mSigma_r \mV_r^\top,
\]
where $\mU_r \in \mathbb{R}^{d \times r}$, $\mSigma_r \in \mathbb{R}^{r \times r}$, and $\mV_r \in \mathbb{R}^{m \times r}$.

\STATE Define the reduced operator
\[
\tilde{\mK} = \mU_r^\top \mZ' \mV_r \mSigma_r^{-1}.
\]

\STATE Compute its eigendecomposition
\[
\tilde{\mK} \vw = \lambda \vw.
\]

\STATE For each eigenpair $(\lambda, \vw)$, define the DMD mode
\[
\vphi = \mU_r \vw.
\]
\end{algorithmic}
\end{algorithm}
Introduced by \citet{schmid_dynamic_2010}, the dynamic mode decomposition (DMD) algorithm describes the dynamics of a given dataset. Given a sequence of states, the algorithm assumes that each state is advanced to the next via a linear model. If we write \(\vx_k\) to represent a state at time \(k\), we can define a sequential dataset with matrices \(\mZ, \mZ' \in \mathbb{R}^{d \times m}\)
\begin{equation}
    \mZ =
    \begin{bmatrix}
        \vline & \vline &        & \vline \\
        \vx_0 & \vx_1 & \cdots & \vx_{m-1} \\
        \vline & \vline &        & \vline
    \end{bmatrix},
    \qquad
    \mZ' =
    \begin{bmatrix}
        \vline & \vline &        & \vline \\
        \vx_1 & \vx_2 & \cdots & \vx_{m} \\
        \vline & \vline &        & \vline
    \end{bmatrix},
\end{equation}
where \(d\) is the number of features and \(m\) is the number of samples. DMD assumes there exists some operator \(\mK \in \mathbb{R}^{d \times d}\), such that
\begin{equation}
\label{eq:lin_dyn_mat}
    \mZ' = \mK\mZ.
\end{equation}
Then, DMD approximates the eigenvalues and eigenvectors of the assumed \(\mK\) for analysis. If the minimization problem is formulated in the Frobenius norm, the least-squares solution is given by the Moore--Penrose pseudoinverse \citep{tu_dynamic_2014}, denoted by \(\mZ^\dagger\):
\begin{equation}
\label{eq:dmd_opt}
    \mK
    = \arg\min_{\mK \in \mathbb{R}^{d \times d}}
    \left\| \mZ' - \mK\mZ \right\|_F^2
    = \mZ'\mZ^\dagger.
\end{equation}
However, \(\mZ\) often has a high ambient dimension, i.e., \(d\) is large. Directly computing, storing, or decomposing \(\mK\) might not be feasible \citep{brunton_modern_2022}. The crux of the DMD algorithm, then, lies in arriving at the eigenvalues and eigenvectors of \(\mK\), \textit{without} explicitly forming the matrix \(\mK\). To achieve this, \citet{schmid_dynamic_2010} proposed \cref{alg:standard_dmd}, which generates a reduced operator \(\tilde{\mK} \in \mathbb{R}^{r\times r}\), which is ultimately eigen-decomposed and its eigenvectors are projected to obtain the \textit{DMD modes} for analysis.

To arrive at a reduced operator \(\tilde{\mK}\), \cref{alg:standard_dmd} compresses the pseudo-inverse solution with \(\mU_r\), the top-\(r\) left singular vectors of \(\mZ\), 
\begin{equation}
\label{eq:dmd_compress}
    \tilde{\mK} = \mU_r^\top \mK \mU_r  = \mU_r^\top \mZ' \mZ^\dag \mU_r.
\end{equation}
In the RHS of \cref{eq:dmd_compress}, \citet{schmid_dynamic_2010} substitutes the pseudoinverse \( \mZ^\dag \approx \mV_r \mSigma_r^{-1} \mU_r^\top\) (obtained from the rank-\(r\) truncated SVD of \(\mZ\)) to arrive at the definition in \cref{alg:standard_dmd}. Subsequent literature \citep{brunton_modern_2022, kostic_learning_2022} has since identified that the standard DMD algorithm is closely tied to principal component regression (PCR), a tool often used in statistics literature but not necessarily applied to time-series data.

\subsection{Equivalence of DMD and PCR}

To clarify the connection between DMD and PCR, we start with the latter. Given \cref{eq:lin_dyn_mat}, we seek a new operator \(\tilde{\mK}\) that satisfies the new regression problem
\begin{equation}
\label{eq:alt_lin_dym_mat}
    \mU_r^\top \mZ' = \tilde{\mK}\mU_r^\top \mZ,
\end{equation}
where instead of regressing the output \(\mZ'\) on the input \(\mZ\), we are regressing the \textit{projected} output \(\mU_r^\top\mZ'\) on the \textit{projected} input \(\mU_r^\top\mZ\). We note that in the standard PCR literature, one only projects the inputs, referred to as the \textit{scores} of \(\mZ\). Solving for \(\tilde{\mK}\) using the normal equations, we arrive at
\begin{equation}
\label{eq:solve_for_K}
    \tilde{\mK} = (\mU_r^\top \mZ')(\mU_r^\top \mZ)^\dag = (\mU_r^\top \mZ')(\mU_r^\top \mU_r \mSigma_r \mV_r^\top)^\dag = \mU_r^\top \mZ' \mV_r \mSigma_r^{-1}
\end{equation}
The RHS of \cref{eq:solve_for_K} is exactly the reduced operator in \cref{alg:standard_dmd}. \citet{kostic_learning_2022} work with another formulation, restating the solutions in terms of covariance matrices. Define
\begin{equation}
\label{eq:cov_defs}
    \hat{\mC} = \mZ\mZ^\top \in \mathbb{R}^{d \times d},
    \qquad
    \hat{\mT} = \mZ'\mZ^\top \in \mathbb{R}^{d \times d},
\end{equation}
the input auto-covariance and the input--output cross-covariance, respectively. In this notation, \cref{eq:dmd_opt} reads $\mK = \hat{\mT}\,\hat{\mC}^\dagger$, and the PCR estimator in \cref{eq:solve_for_K} becomes
\begin{equation}
\label{eq:pcr_cov}
    \mK = \hat{\mT}\,[[\hat{\mC}]]_r^\dagger,
\end{equation}
where $[[\cdot]]_r$ denotes the best rank-$r$ approximation. That is, PCR truncates the input covariance to rank $r$, retaining only the top-$r$ eigenvectors of $\hat{\mC}$ (equivalently, the top-$r$ left singular vectors of $\mZ$). Compressing \cref{eq:pcr_cov} to the $r$-dimensional subspace via $\mU_r$ recovers the reduced operator of \cref{eq:solve_for_K}:
\begin{equation}
\label{eq:pcr_compress}
    \mU_r^\top\hat{\mT}\,[[\hat{\mC}]]_r^\dagger\,\mU_r
    = \mU_r^\top\mZ'\mV_r\mSigma_r^{-1}
    = \tilde{\mK}.
\end{equation}

The covariance formulation makes regularization a natural modification \citep{kostic_learning_2022}: for $\alpha > 0$, define $\hat{\mC}_\alpha = \hat{\mC} + \alpha\mI$ and consider the ridge estimator
\begin{equation}
\label{eq:ridge}
    \mK = \hat{\mT}\,\hat{\mC}_\alpha^{-1}.
\end{equation}
This replaces the pseudoinverse with a regularized inverse, stabilizing the estimate without imposing a rank constraint.

\subsection{Reduced Rank Regression}
\label{app:rrr}

PCR selects its rank-$r$ subspace by truncating $\hat{\mC}$, retaining directions of greatest input variance. However, this does not minimize the prediction error under a rank constraint. Directions of high input variance are not necessarily the most predictive of $\mZ'$ \citep{brunton_modern_2022, kostic_learning_2022}. Proposed by \citet{kostic_learning_2022}, reduced rank regression (RRR) combines ridge regularization with rank truncation, selecting directions that maximize predictive power instead. The RRR estimator of rank~$r$ with regularization $\alpha > 0$ solves

\begin{equation}
\label{eq:rrr_obj}
    \mK_{r,\alpha}^{\mathrm{RRR}}
    = \argmin_{\substack{\mK \in \mathbb{R}^{d \times d}\\
      \mathrm{rank}(\mK) \leq r}}
    \left\|\mZ' - \mK\mZ\right\|_F^2
    + \alpha\left\|\mK\right\|_F^2,
\end{equation}

with closed-form solution

\begin{equation}
\label{eq:rrr_sol}
    \mK_{r,\alpha}^{\mathrm{RRR}}
    = \left[\!\left[
        \hat{\mT}\,\hat{\mC}_\alpha^{-1/2}
      \right]\!\right]_r
      \hat{\mC}_\alpha^{-1/2}.
\end{equation}

As in \cref{eq:ridge}, the penalty $\alpha\|\mK\|_F^2$ shifts the input covariance to $\hat{\mC}_\alpha = \hat{\mC} + \alpha\mI$. The difference from PCR lies in where the rank-$r$ truncation is applied; PCR truncates $\hat{\mC}$ (the input covariance), while RRR truncates $\hat{\mT}\,\hat{\mC}_\alpha^{-1/2}$ (the whitened cross-covariance). The retained directions therefore maximize predictive power rather than input variance. 

\begin{figure}[b!]
    \centering
    \includegraphics[width=\linewidth]{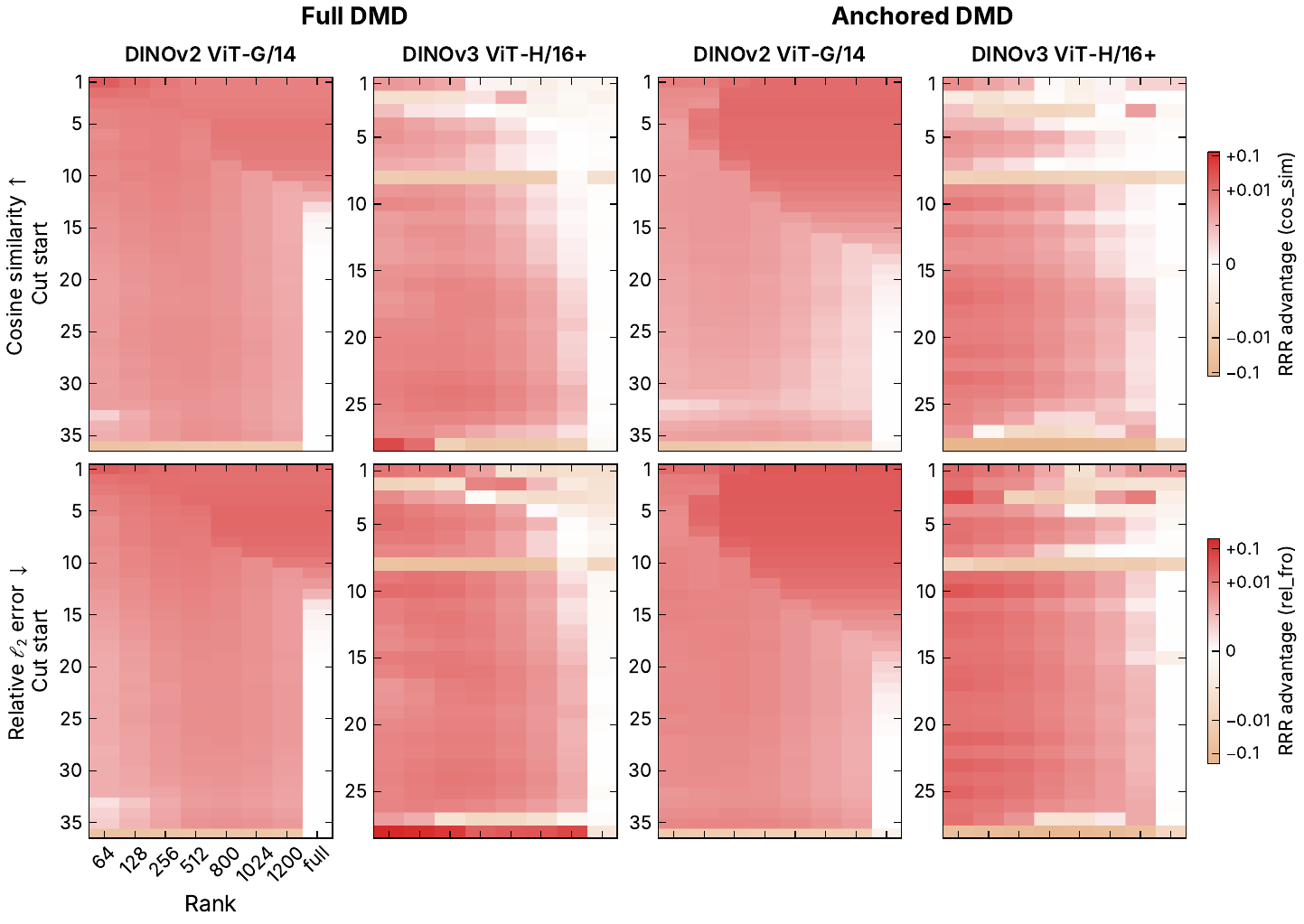}
    \caption{\textbf{PCR vs.\ RRR.} Per-cell difference (RRR minus PCR) in cosine similarity (top) and relative \(\ell_2\) error (bottom) across operator ranks, cut starts, and DINO models under \fullDMD and \anchoredDMD. Symmetric log color scale; red indicates RRR ahead, peach indicates PCR ahead. The rightmost column corresponds to full rank, where the two solvers coincide mathematically.}
    \label{fig:pcr_vs_rrr}
\end{figure}

We emphasize that both algorithms solve the same problem with slightly different approaches. Our experiments use the \texttt{Ridge} class in the \texttt{kooplearn} library \citep{turri_kooplearn_2026}, which supports both PCR and RRR, governed by the truncation rank $r$. \cref{fig:pcr_vs_rrr} plots the per-cell gap between RRR and PCR (RRR minus PCR) across operator ranks, cut starts, and four DINO models, on cosine similarity (top) and relative \(\ell_2\) error (bottom). 

The picture is broadly red, indicating that RRR has a slight overall edge across the grid. Note that the color scale is symmetric log, so even modest red shading corresponds to differences below \(\pm 0.01\); the magnitudes are small and the choice between solvers rarely matters operationally. The rightmost column (full rank) is essentially white throughout, consistent with PCR and RRR coinciding mathematically at full rank. The residual non-zero cells in this column reflect numerical implementation differences in the two solvers. Outside the full-rank column, RRR's edge is consistent but small. A few configurations break the pattern. At the deepest cut on DINOv3-H/16+ (anchored, cs=28) and DINOv2-G/14 (full, cs=35), RRR underperforms PCR uniformly across ranks. On DINOv3-H/16+ full DMD at cs=8, the two metrics disagree, with PCR ahead on cosine similarity and RRR ahead on relative \(\ell_2\) error. We default to RRR throughout the paper, but the analysis would survive substituting PCR with negligible change in the headline numbers.

\section{Calibration Budget}
\label{app:calib}

\Cref{fig:calib_gamma} presents a power-law analysis of calibration convergence. For each configuration (model, cut-start position, DMD formulation), we normalize the relative \(\ell_2\) error at calibration budget \(B\) by its value at \(B{=}1{,}000\) and fit the excess \(\ell_2(B)\,/\,\ell_2(1000) - 1\) to a power law \(C/B^\gamma\) using \texttt{scipy.curve\_fit}, which returns \(C\) and \(\gamma\). The fitted \(\gamma\) cluster near \(\approx 1\), indicating an \(O(1/B)\) convergence. In practice, both \fullDMD and \anchoredDMD converge to within 5\% of their large-budget baseline by \(B = 250\) calibration images. 

\begin{figure}[htbp!]
    \centering
    \includegraphics[width=\linewidth]{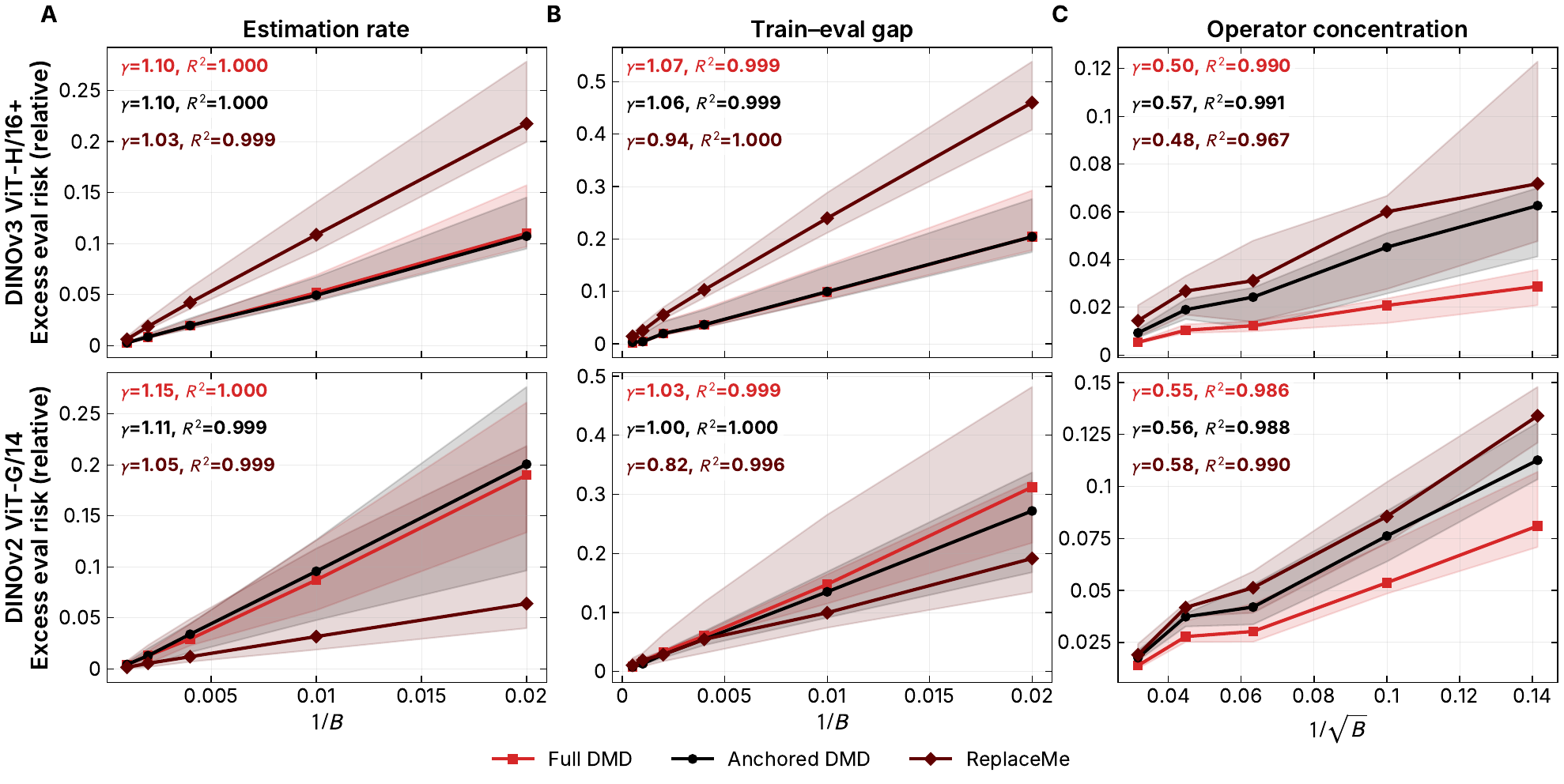}
    \caption{\textbf{Empirical validation of calibration budget learning bounds \citep{kostic_learning_2022}} (\(\alpha = 10^{-4}\))\textbf{.} DINOv3 ViT-H/16+ (top) and DINOv2 ViT-G/14 (bottom). \textbf{(A)} Evaluation risk ratio vs.\ $1/B$. \textbf{(B)} Train-eval MSE gap vs.\ $1/B$. \textbf{(C)} Cross-covariance concentration vs.\ $1/\sqrt{B}$.}
    \label{fig:theory_triptych}
\end{figure}
\vspace{-2em}
\begin{table}[h]
\centering
\caption{Median empirical constant $C_{\mathrm{emp}}$ from fitting $1 + C/B^\gamma$ to the evaluation risk ratio at prune length $p=5$, compared to four predicted constants.
Ratios near 1 indicate agreement. The reported ratio is the median of per-configuration ratios $C_{\mathrm{emp},i} / C_{\mathrm{pred},i}$ across cut starts, not the ratio of the displayed medians.} \label{tab:C_comparison}
\scriptsize
\begin{tabular}{ll r cccc}
\toprule
Model & Method & $C_{\mathrm{emp}}$ & $d/(tp)$ & $d/t$ & $d^2/(tp)$ & $d^2/t$ \\
\cmidrule(lr){4-4}\cmidrule(lr){5-5}\cmidrule(lr){6-6}\cmidrule(lr){7-7}
& & & $C_{\mathrm{pred}}$ / ratio & $C_{\mathrm{pred}}$ / ratio & $C_{\mathrm{pred}}$ / ratio & $C_{\mathrm{pred}}$ / ratio \\ \midrule
\multirow{2}{*}{\shortstack[l]{DINOv3\\[-2pt]\scriptsize $d{=}1280,\; t{=}197$}} & Full     & 9.2  & 1.30 / 7.4  & 6.50 / 1.4 & $1.7{\times}10^3$ / $5.8{\times}10^{-3}$ & $8.3{\times}10^3$ / $1.1{\times}10^{-3}$ \\ & Anchored & 8.2  & 1.30 / 7.0  & 6.50 / 1.3 & $1.7{\times}10^3$ / $5.5{\times}10^{-3}$ & $8.3{\times}10^3$ / $9.9{\times}10^{-4}$ \\ \midrule
\multirow{2}{*}{\shortstack[l]{DINOv2\\[-2pt]\scriptsize $d{=}1536,\; t{=}257$}} & Full     & 19.3 & 1.20 / 14.9 & 5.98 / 3.2 & $1.8{\times}10^3$ / $9.7{\times}10^{-3}$ & $9.2{\times}10^3$ / $2.1{\times}10^{-3}$ \\ & Anchored & 11.5 & 1.20 / 9.8  & 5.98 / 1.9 & $1.8{\times}10^3$ / $6.4{\times}10^{-3}$ & $9.2{\times}10^3$ / $1.3{\times}10^{-3}$ \\ \bottomrule
\end{tabular}
\end{table}

We now extend this to validate findings from \citet{kostic_learning_2022} (\cref{fig:theory_triptych}), using a different set of 15 configurations and including ReplaceMe. Panel A plots the evaluation risk ratio (MSE at \(B\) normalized by MSE at \(B{=}2{,}000\), a different reference and presentation than the excess fit above) against \(1/B\). Fitting this ratio to \(1 + C/B^\gamma\) confirms \(\gamma \approx 1\) across all three methods. \(\gamma\) alone does not fully characterize the rate, as both \(d/(tB)\) and \(d^2/(tB)\) could be valid; the distinction lies in the constant \(C\). Recall that \(M = B \cdot p \cdot t\) is the raw sample count when fitting DMD. Then, does \(p\) truly inflate the effective sample count? To test this, we consider four candidates \(C\):
\begin{equation}
C_{d/tp} = \frac{d}{tp}, \qquad
C_{d/t} = \frac{d}{t}, \qquad
C_{d^2/tp} = \frac{d^2}{tp}, \qquad
C_{d^2/t} = \frac{d^2}{t},
\end{equation}
where \(t\) is the number of tokens per image, \(p\) is the prune length, and \(d\) is the feature dimension. \Cref{tab:C_comparison} compares \(C_{\mathrm{emp}}\) against each candidate. The \(d^2\) candidates overestimate \(C\) by three orders of magnitude. Among the \(d\)-family, the \(d/t\) model yields ratios within roughly $3\times$ across both DMD variants and both models. This is consistent with the $\gamma\approx 1$ rate but does not pin the constant tightly; the ordering nonetheless rules out the \(d/(tp)\) candidate, suggesting that increasing \(p\) does not proportionally inflate the effective sample count. 

\begin{figure}[b!]
    \centering
    \includegraphics[width=\linewidth]{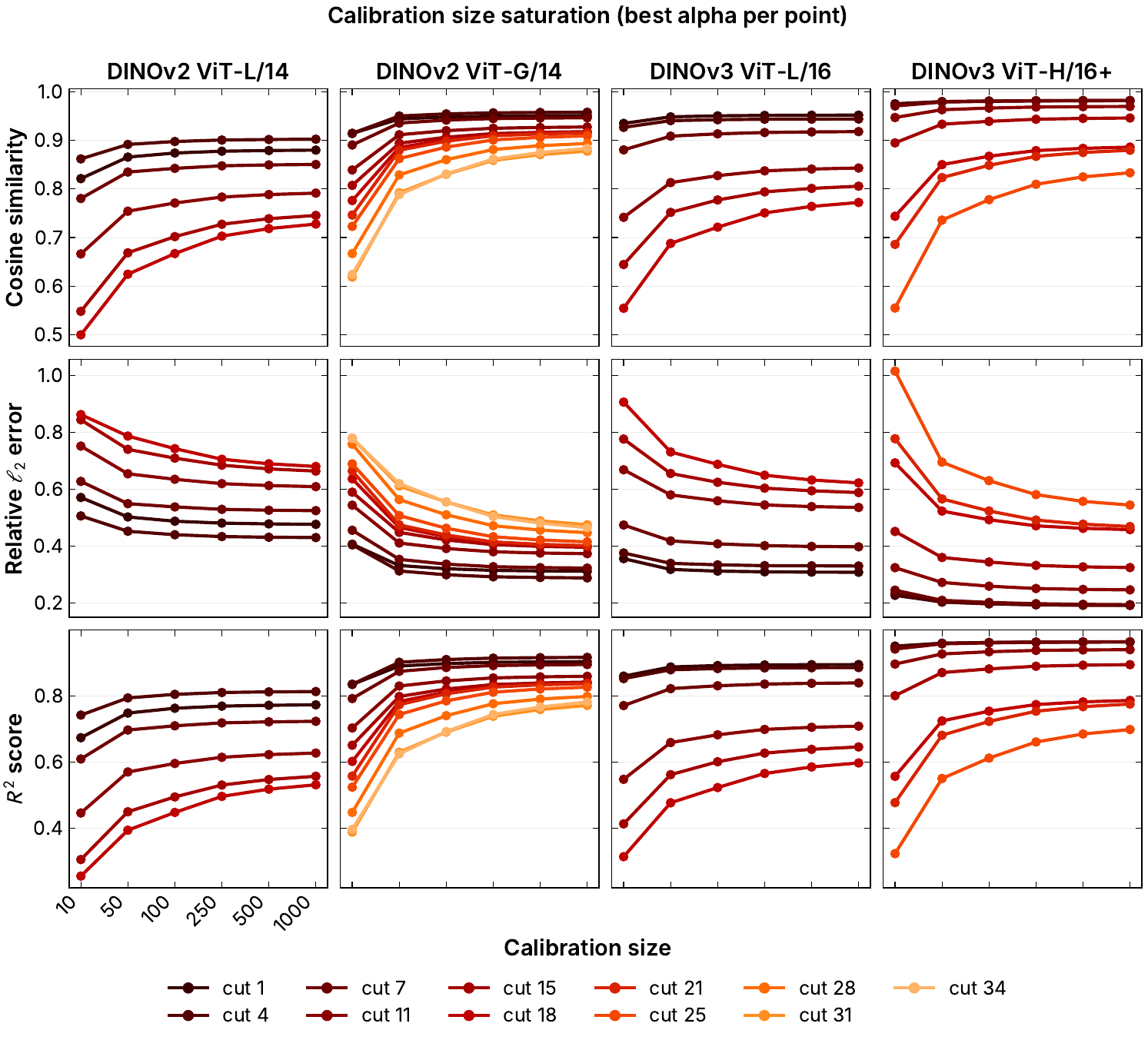}
    \caption{\textbf{Calibration budget saturation under \fullDMD (RRR, \(\alpha = 10^{-5}\)).} Cosine similarity (top), relative \(\ell_2\) error (second), \(R^2\) score (third), and norm ratio (bottom) at prediction step \(q{=}3\) as a function of calibration set size \(B \in \{10, 50, 100, 250, 500, 1000\}\). Each line corresponds to a cut start. Flat lines indicate that a small calibration set suffices; steeply rising lines indicate that the operator fit is data-hungry at that depth. For DINOv3-L/16, we omit cut start \(i=1\) due to an ill-conditioned boundary fit at this \(\alpha\) (relative \(\ell_2 > 3\) at small \(B\)).}
    \label{fig:calib_saturation}
\end{figure}

\Cref{fig:theory_triptych}B plots the train and evaluation MSE gap, normalized by the evaluation MSE at \(B{=}2{,}000\), against \(1/B\). Theorem~3 of \citet{kostic_learning_2022} bounds this gap at \(O(1/\sqrt{B})\); the DMD methods show \(\gamma \approx 1\), indicating \(O(1/B)\) decay, faster than the bound predicts. For both models, ReplaceMe decays more slowly (\(\gamma = 0.82\)--\(0.94\)). Finally, panel C plots cross-covariance concentration against \(1/\sqrt{B}\). All three methods yield \(\gamma \approx 0.5\), confirming the \(O(1/\sqrt{B})\) rate of Proposition~2 \citep{kostic_learning_2022}.

In \cref{fig:calib_saturation}, we also present per-model saturation curves across four DINO architectures, which explains the range in \cref{fig:calib_gamma}B. The figure shows that early cuts saturate quickly across all models, with gains beyond $B = 250$ being modest. Late cuts are universally data-hungry, showing continued improvement through $B = 1{,}000$ across all architectures.

\section{DMD Formulation}
\label{app:fusability}

\begin{figure}[hb!]
    \centering
    \includegraphics[width=\linewidth]{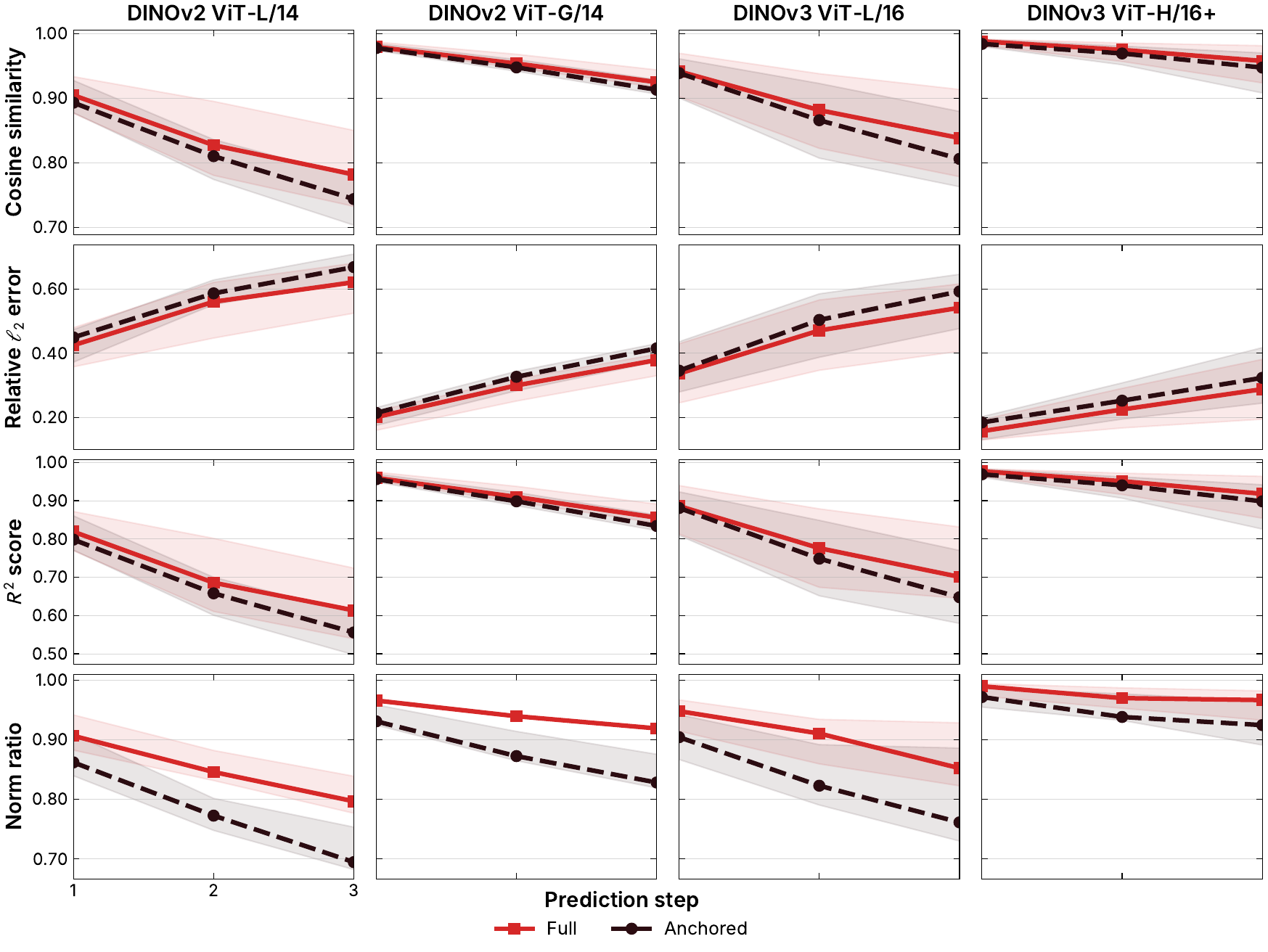}
    \caption{\textbf{\FullDMD vs.\ \anchoredDMD approximation quality over prediction depth.} Cosine similarity, relative $\ell_2$ error, and $R^2$ score between predicted and true hidden states at prediction steps $q \in \{1,2,3\}$, across all four DINO variants. Lines show the mean across cut starts $i \in \{2,\dots,15\}$; shaded regions show the per-cut min/max. One-step operators are fit as full-rank unregularized linear maps ($\alpha = 0$).}
    \label{fig:fusability}
\end{figure}

The two formulations, \fullDMD and \anchoredDMD, are defined in \cref{sec:method}. \Cref{fig:fusability} shows that the two approaches perform similarly across all four architectures. \FullDMD holds a slight advantage on relative $\ell_2$ error and $R^2$, but the difference is modest and both formulations are highly variable across cut starts, as reflected in the wide shaded bands. This variability is especially pronounced in the L variants (DINOv2 ViT-L/14 and DINOv3 ViT-L/16), where approximation quality varies sharply with cut depth.
\FloatBarrier

\section{Experiment Details}
\label{app:sweep_details}

\begin{figure}[htb!]
    \centering
    \makebox[\textwidth][c]{\includegraphics[width=1.4\textwidth]{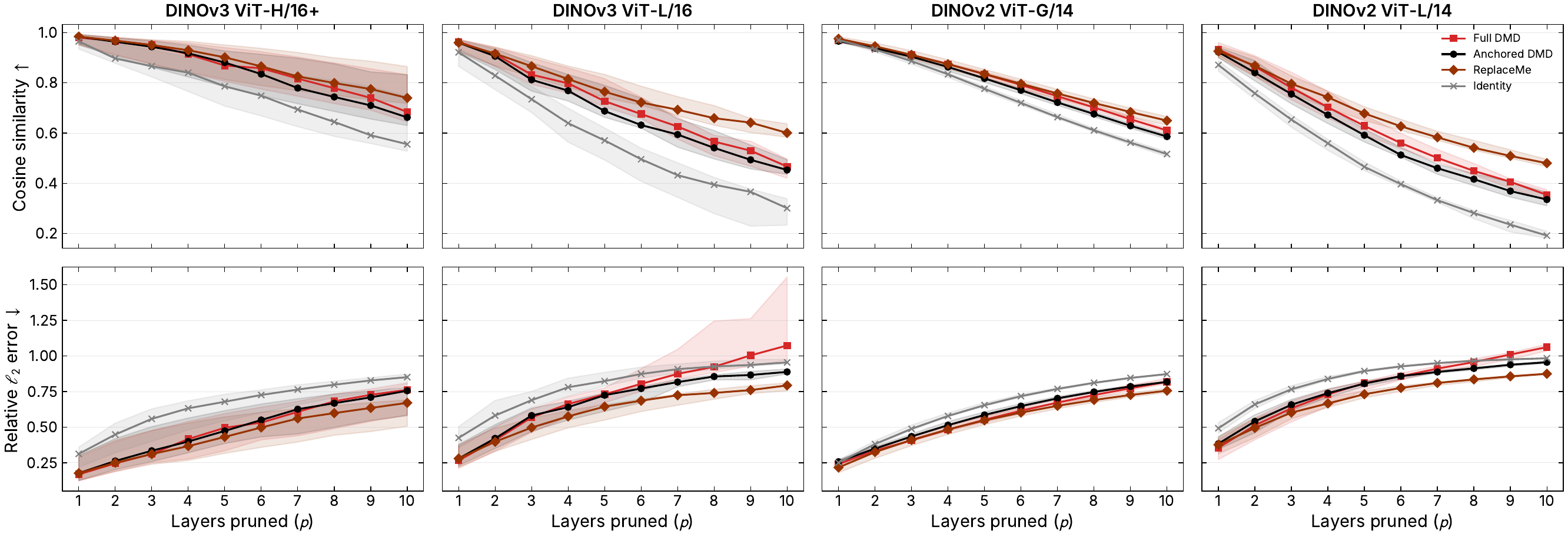}}
    \caption{\textbf{Reconstruction quality across prune lengths, all four DINO models.} Extends \cref{fig:headline} to the DINOv2-L/14 and DINOv3-L/16 variants. Per-position breakdown of the metrics in \cref{tab:headline_full}, across all prune lengths $p$.}
    \label{fig:headline_full}
\end{figure}

In \cref{tab:headline_full} and \cref{fig:headline_full}, we extend the sweep results from \cref{tab:headline} and \cref{fig:headline}, including two additional DINO models.

\FloatBarrier

\begin{table*}[t]
\centering
\caption{Cosine similarity\,$\uparrow$\;/\;relative $\ell_2$ error\,$\downarrow$ of the predicted hidden state across all four DINO models, reported as the median over all cut-start positions.}
\label{tab:headline_full}
\footnotesize
\makebox[\textwidth][c]{%
\begin{tabular}{l c c c c c c c c c c}
\toprule
Method & $p=1$ & $p=2$ & $p=3$ & $p=4$ & $p=5$ & $p=6$ & $p=7$ & $p=8$ & $p=9$ & $p=10$ \\
\midrule
\multicolumn{11}{l}{\textbf{DINOv3 ViT-H/16+}} \\
Full DMD & 0.98\,/\,0.17 & 0.97\,/\,0.25 & 0.95\,/\,0.31 & 0.91\,/\,0.42 & 0.87\,/\,0.50 & 0.86\,/\,0.53 & 0.82\,/\,0.61 & 0.78\,/\,0.68 & 0.74\,/\,0.73 & 0.68\,/\,0.76 \\
Anchored DMD & 0.98\,/\,0.18 & 0.96\,/\,0.26 & 0.94\,/\,0.33 & 0.92\,/\,0.40 & 0.88\,/\,0.47 & 0.84\,/\,0.55 & 0.78\,/\,0.63 & 0.74\,/\,0.67 & 0.71\,/\,0.71 & 0.66\,/\,0.76 \\
ReplaceMe & 0.98\,/\,0.18 & 0.97\,/\,0.25 & 0.95\,/\,0.31 & 0.93\,/\,0.37 & 0.90\,/\,0.43 & 0.87\,/\,0.50 & 0.83\,/\,0.56 & 0.80\,/\,0.60 & 0.78\,/\,0.63 & 0.74\,/\,0.67 \\
Identity & 0.97\,/\,0.31 & 0.90\,/\,0.45 & 0.87\,/\,0.56 & 0.84\,/\,0.63 & 0.79\,/\,0.68 & 0.75\,/\,0.73 & 0.69\,/\,0.76 & 0.64\,/\,0.80 & 0.59\,/\,0.83 & 0.56\,/\,0.85 \\
\addlinespace
\multicolumn{11}{l}{\textbf{DINOv3 ViT-L/16}} \\
Full DMD & 0.96\,/\,0.27 & 0.91\,/\,0.41 & 0.83\,/\,0.57 & 0.80\,/\,0.66 & 0.73\,/\,0.73 & 0.68\,/\,0.80 & 0.63\,/\,0.87 & 0.57\,/\,0.92 & 0.53\,/\,1.00 & 0.47\,/\,1.07 \\
Anchored DMD & 0.96\,/\,0.28 & 0.91\,/\,0.42 & 0.81\,/\,0.58 & 0.77\,/\,0.64 & 0.69\,/\,0.72 & 0.63\,/\,0.77 & 0.59\,/\,0.82 & 0.54\,/\,0.86 & 0.49\,/\,0.87 & 0.45\,/\,0.89 \\
ReplaceMe & 0.96\,/\,0.28 & 0.92\,/\,0.40 & 0.87\,/\,0.50 & 0.82\,/\,0.58 & 0.76\,/\,0.64 & 0.72\,/\,0.69 & 0.69\,/\,0.72 & 0.66\,/\,0.74 & 0.64\,/\,0.76 & 0.60\,/\,0.79 \\
Identity & 0.92\,/\,0.42 & 0.83\,/\,0.58 & 0.73\,/\,0.69 & 0.64\,/\,0.78 & 0.57\,/\,0.82 & 0.50\,/\,0.87 & 0.43\,/\,0.91 & 0.39\,/\,0.93 & 0.37\,/\,0.94 & 0.30\,/\,0.95 \\
\addlinespace
\multicolumn{11}{l}{\textbf{DINOv2 ViT-G/14}} \\
Full DMD & 0.97\,/\,0.24 & 0.94\,/\,0.33 & 0.91\,/\,0.41 & 0.87\,/\,0.48 & 0.83\,/\,0.55 & 0.79\,/\,0.62 & 0.75\,/\,0.67 & 0.70\,/\,0.73 & 0.66\,/\,0.77 & 0.61\,/\,0.82 \\
Anchored DMD & 0.97\,/\,0.26 & 0.94\,/\,0.35 & 0.90\,/\,0.43 & 0.86\,/\,0.51 & 0.82\,/\,0.59 & 0.77\,/\,0.65 & 0.72\,/\,0.70 & 0.68\,/\,0.75 & 0.63\,/\,0.79 & 0.59\,/\,0.82 \\
ReplaceMe & 0.98\,/\,0.22 & 0.95\,/\,0.32 & 0.91\,/\,0.41 & 0.87\,/\,0.48 & 0.84\,/\,0.55 & 0.80\,/\,0.60 & 0.76\,/\,0.65 & 0.72\,/\,0.69 & 0.68\,/\,0.73 & 0.65\,/\,0.76 \\
Identity & 0.97\,/\,0.25 & 0.93\,/\,0.38 & 0.89\,/\,0.49 & 0.83\,/\,0.58 & 0.78\,/\,0.65 & 0.72\,/\,0.72 & 0.66\,/\,0.77 & 0.61\,/\,0.81 & 0.56\,/\,0.85 & 0.52\,/\,0.87 \\
\addlinespace
\multicolumn{11}{l}{\textbf{DINOv2 ViT-L/14}} \\
Full DMD & 0.93\,/\,0.36 & 0.87\,/\,0.51 & 0.78\,/\,0.63 & 0.70\,/\,0.73 & 0.63\,/\,0.81 & 0.56\,/\,0.86 & 0.50\,/\,0.91 & 0.45\,/\,0.96 & 0.41\,/\,1.01 & 0.35\,/\,1.06 \\
Anchored DMD & 0.92\,/\,0.38 & 0.84\,/\,0.54 & 0.76\,/\,0.66 & 0.67\,/\,0.74 & 0.59\,/\,0.81 & 0.51\,/\,0.86 & 0.46\,/\,0.89 & 0.42\,/\,0.91 & 0.37\,/\,0.94 & 0.34\,/\,0.96 \\
ReplaceMe & 0.93\,/\,0.38 & 0.87\,/\,0.49 & 0.80\,/\,0.60 & 0.74\,/\,0.67 & 0.68\,/\,0.73 & 0.63\,/\,0.78 & 0.58\,/\,0.81 & 0.54\,/\,0.84 & 0.51\,/\,0.86 & 0.48\,/\,0.87 \\
Identity & 0.87\,/\,0.49 & 0.76\,/\,0.66 & 0.65\,/\,0.77 & 0.56\,/\,0.84 & 0.47\,/\,0.89 & 0.40\,/\,0.93 & 0.33\,/\,0.95 & 0.28\,/\,0.97 & 0.24\,/\,0.98 & 0.19\,/\,0.98 \\
\bottomrule
\end{tabular}}
\end{table*}

We verify that the method orderings observed in \cref{tab:headline} and \cref{tab:headline_full} are statistically significant using a Friedman test with Nemenyi post-hoc \citep{demsar_statistical_2006}. \Cref{tab:friedman} reports the average rank of each method across all (cut-start, prune-length) configurations per model. The Friedman \(p < 10^{-69}\) in every model and metric, indicating that the four methods are not equivalent. ReplaceMe ranks first in every model on both metrics. Among the DMD variants, \fullDMD beats \anchoredDMD on cosine similarity in every model, with all rank gaps exceeding the critical difference; on relative \(\ell_2\) error, \fullDMD leads on DINOv3-H/16+ and DINOv2-G/14, \anchoredDMD on DINOv3-L/16, and the two tie on DINOv2-L/14. The identity baseline ranks last on both metrics in every model, well below all three learned methods. We note that adjacent (cut-start, prune-length) configurations share evaluation data and overlapping spans, so the independence assumption is not strictly satisfied; we therefore read this test as a coarse ordering check rather than a formal significance claim.

\begin{table*}[t]
 \centering
 \caption{Friedman test with Nemenyi post-hoc \citep{demsar_statistical_2006}. Average ranks across all (cut-start, prune-length) configurations per model, lower is better. Friedman $p < 10^{-69}$ in every cell. Critical difference (CD) for Nemenyi at $\alpha{=}0.05$ is reported per model; pairs of methods whose rank gap exceeds CD differ significantly.}
 \label{tab:friedman}
 \small
 \begin{tabular}{l c c c c c c c c}
 \toprule
 & \multicolumn{2}{c}{DINOv3 ViT-H/16+} & \multicolumn{2}{c}{DINOv2 ViT-G/14} & \multicolumn{2}{c}{DINOv3 ViT-L/16} & \multicolumn{2}{c}{DINOv2 ViT-L/14} \\
 & \multicolumn{2}{c}{$n{=}245$, CD\,=\,$0.300$} & \multicolumn{2}{c}{$n{=}325$, CD\,=\,$0.260$} & \multicolumn{2}{c}{$n{=}165$, CD\,=\,$0.365$} & \multicolumn{2}{c}{$n{=}165$, CD\,=\,$0.365$} \\
 \cmidrule(lr){2-3} \cmidrule(lr){4-5} \cmidrule(lr){6-7} \cmidrule(lr){8-9}
 Method & Cos\,$\downarrow$ & Rel\,$\ell_2$\,$\downarrow$ & Cos\,$\downarrow$ & Rel\,$\ell_2$\,$\downarrow$ & Cos\,$\downarrow$ & Rel\,$\ell_2$\,$\downarrow$ & Cos\,$\downarrow$ & Rel\,$\ell_2$\,$\downarrow$ \\
 \midrule
 Full DMD     & 1.93 & 1.95 & 1.83 & 1.98 & 2.07 & 2.77 & 1.82 & 2.29 \\
 Anchored DMD & 2.86 & 2.80 & 3.03 & 2.93 & 2.78 & 2.36 & 2.95 & 2.62 \\
 ReplaceMe    & 1.27 & 1.31 & 1.28 & 1.23 & 1.16 & 1.14 & 1.25 & 1.28 \\
 Identity     & 3.95 & 3.95 & 3.85 & 3.86 & 3.99 & 3.73 & 3.98 & 3.81 \\
 \bottomrule
 \end{tabular}
 \end{table*}

\newpage
\section{Additional Experiments}
\label{app:additional_experiments}

\begin{figure}[htb!]
    \centering
    \includegraphics[width=0.9\textwidth]{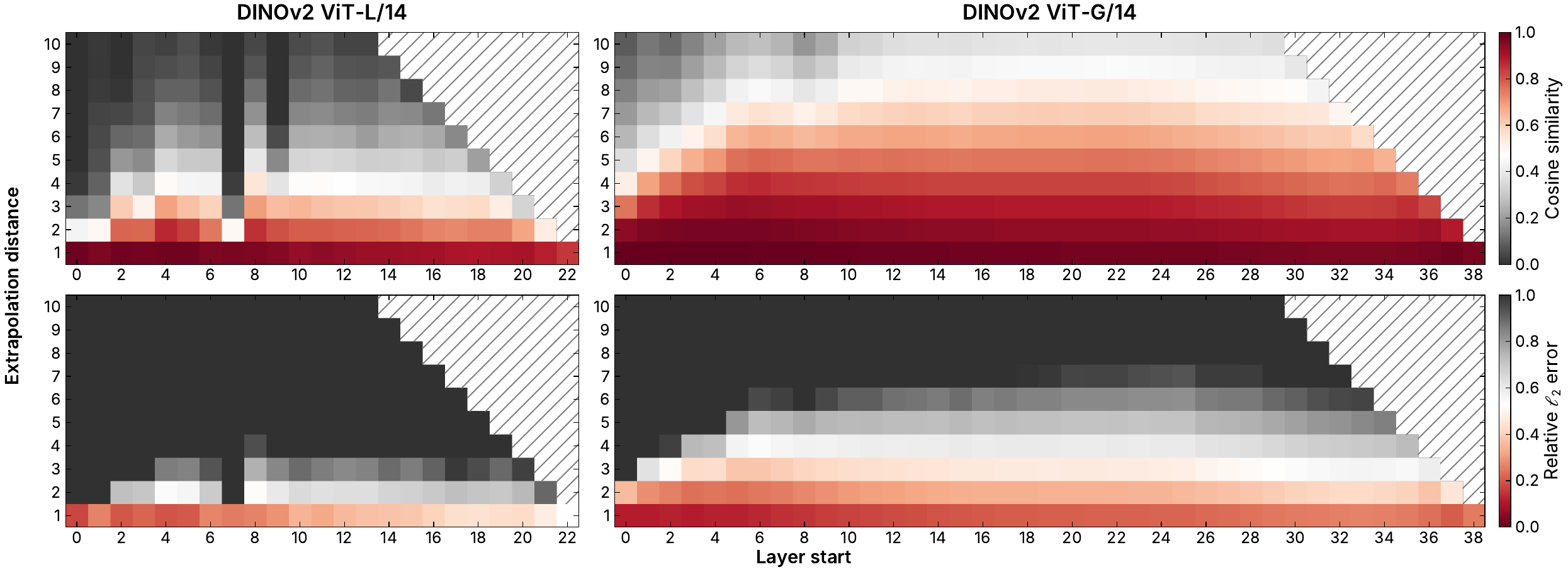}
    \vspace{0.5em}
    \includegraphics[width=0.9\textwidth]{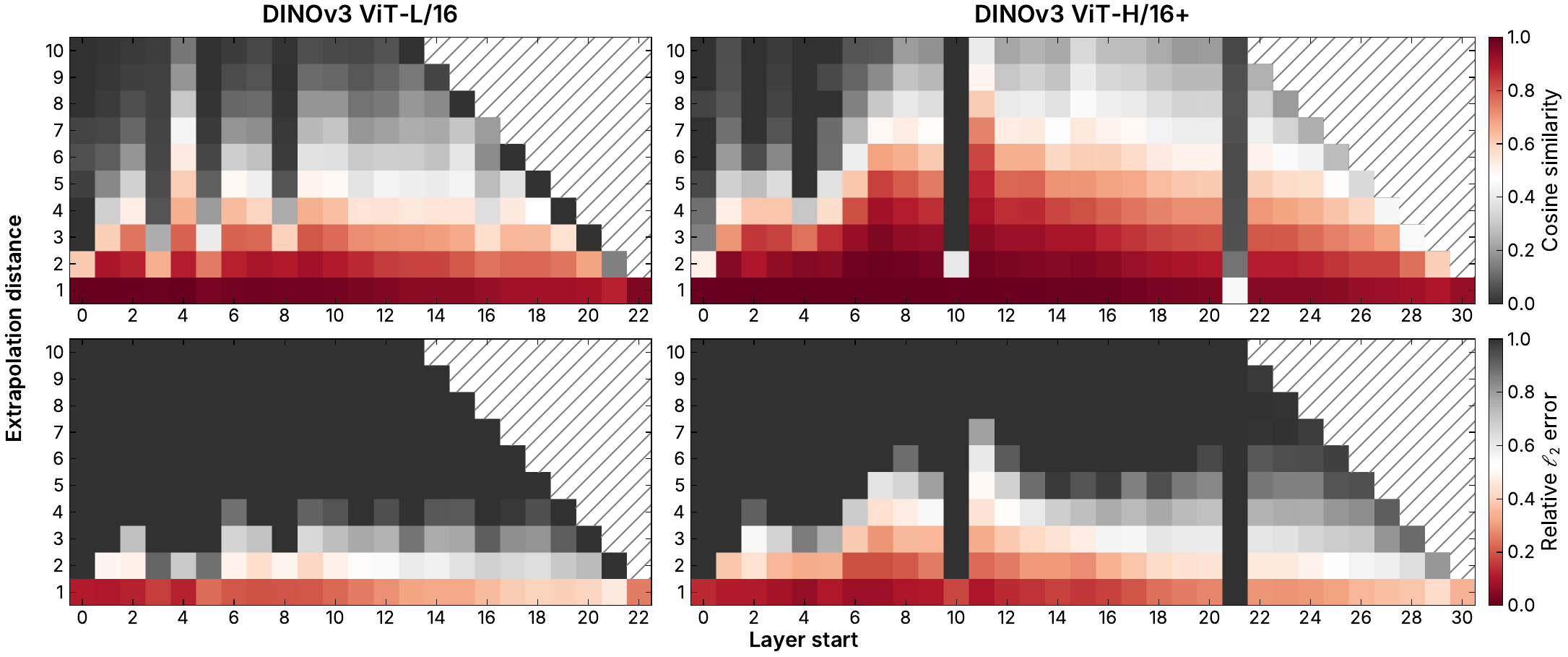}
    \caption{\textbf{Multi-step extrapolation across model families.} Same protocol as \cref{fig:extrap} applied to DINOv2 (top; ViT-L/14 and ViT-G/14) and DINOv3 (bottom; ViT-L/16 and ViT-H/16+).}
    \label{fig:extrap-all}
\end{figure}







\end{document}